\documentclass[journal,12pt,onecolumn,draftclsnofoot,]{IEEEtran}
\usepackage{cite}
\usepackage{amsmath,amssymb,amsfonts}
\usepackage{algorithmic}
\usepackage{graphicx}
\usepackage{textcomp}
\usepackage{color,soul}
\usepackage{subcaption}
\usepackage{url}
\usepackage{dirtytalk}
\DeclareUnicodeCharacter{2061}{}

\begin{document}

\title{Wireless End-to-End Image Transmission System using Semantic Communications}
\author{Maheshi~Lokumarambage,~\IEEEmembership{Student~Member, IEEE,} Vishnu~Gowrisetty,~\IEEEmembership{Student~Member, IEEE,} Hossein~Rezaei,~\IEEEmembership{Student~Member, IEEE,} Thushan~Sivalingam,~\IEEEmembership{Student Member, IEEE,} Nandana~Rajatheva,~\IEEEmembership{Senior~Member, IEEE,} Anil~Fernando,~\IEEEmembership{Senior~Member, IEEE}}
\maketitle
\begin{abstract}
Semantic communication is considered the future of mobile communication, which aims to transmit data beyond Shannon's theorem of communications by transmitting the semantic meaning of the data rather than the bit-by-bit reconstruction of the data at the receiver's end. The semantic communication paradigm aims to bridge the gap of limited bandwidth problems in modern high-volume multimedia application content transmission. Integrating AI technologies with the 6G communications networks paved the way to develop semantic communication-based end-to-end communication systems. In this study, we have implemented a semantic communication-based end-to-end image transmission system, and we discuss potential design considerations in developing semantic communication systems in conjunction with physical channel characteristics. A Pre-trained GAN network is used at the receiver as the transmission task to reconstruct the realistic image based on the Semantic segmented image at the receiver input. The semantic segmentation task at the transmitter (encoder) and the GAN network at the receiver (decoder) is trained on a common knowledge base, the COCO-Stuff dataset. The research shows that the resource gain in the form of bandwidth saving is immense when transmitting the semantic segmentation map through the physical channel instead of the ground truth image in contrast to conventional communication systems. Furthermore, the research studies the effect of physical channel distortions and quantization noise on semantic communication-based multimedia content transmission.    
\end{abstract}


\maketitle
\section{Introduction}

A key emerging challenge with the surge in multimedia traffic and the ever increasing use of wireless sensor networks (WSN) and Internet of Things (IoT) components are causing several sustainability problems in managing communication netoworks. The increasingly complex nature of media and communication systems has led to an enormous increase in its bandwidth and energy demands. Sustainability needs to be applied pervasively across such systems in order to bring a significant overall reduction for the negative impact of the resource utilisation and environment. A key requirement is the minimization of the bandwidth and energy footprint of media transmissions, WSNs and IoTs using different network technologies (such as various IEEE 802 standards, as well as 5G and 6G systems) together with reduction of the ever-increasing traffic load due to multimedia traffic. These services and networks are also supposed to support smart and adaptive operation using complex control frameworks, which even increases resource consumption further. 
Semantic communication(SC) is a paradigm which has renewed academic and industry interest, mainly due to the promising possibilities it allows to go beyond Shannon’s capacity limit in bandwidth limited communication channels. The objective is to deliver the semantic meaning of the message and not the exact form of the message by sharing a common, prior knowledge and a semantically encoded message, which is expected to perform better than state of the art compression techniques, thereby drastically reducing the physical bandwidth requirement between the transmitter and receiver. While the applications and expected benefits are obvious in high-bandwidth applications such as super high resolution (16K video at higher frame rates) video transmission, 3D video, AR/VR/MR streaming, the trade off of such approaches regarding effort versus gains is less clear for machine-to-machine (M2M) communication, WSN and IoT. Yet, overall with Semantic communications concepts, it is expected to reduce the bandwidth, complexity, increase the range and enable longer operational cycles in battery powered devices for WSNs and IoTs and M2M communications.

The traditional communications paradigm is focused on transmitting a minimum number of bits with the smallest possible errors between two points. This was based on Shannon's original paper in 1948~\cite{shannon_mathematical_1949} on establishing channel capacity and proof that all rates below capacity are possible without incurring an exponentially higher number of errors at the receiver side. Research continued for over half a century before capacity-achieving codes at long block lengths were found. The receiver does not exploit the information about the source available at the transmitter side explicitly. While there has been a lot of research on source and channel coding, as well as unequal error protection (UEP), this was essentially still a matter on the transmitter side. Further, they suggested that there are three levels in a communication system, each with a specific task: the technical problem, the semantic problem, and the effectiveness problem. The technical problem pertains to how effectively the symbols of the message are transmitted; the semantic problem pertains to how effectively the transmitted symbols convey the meaning of the message intended to be transmitted, and the effectiveness problem pertains to how effectively the transmitted symbols are doing the intended task at the receiver. Traditional communication systems have mainly focused on addressing the technical problem. In semantic communications, it is planned to address the second layer which is the semantic problem.

\begin{figure*}[t]
    \center
    \includegraphics[width=\linewidth]{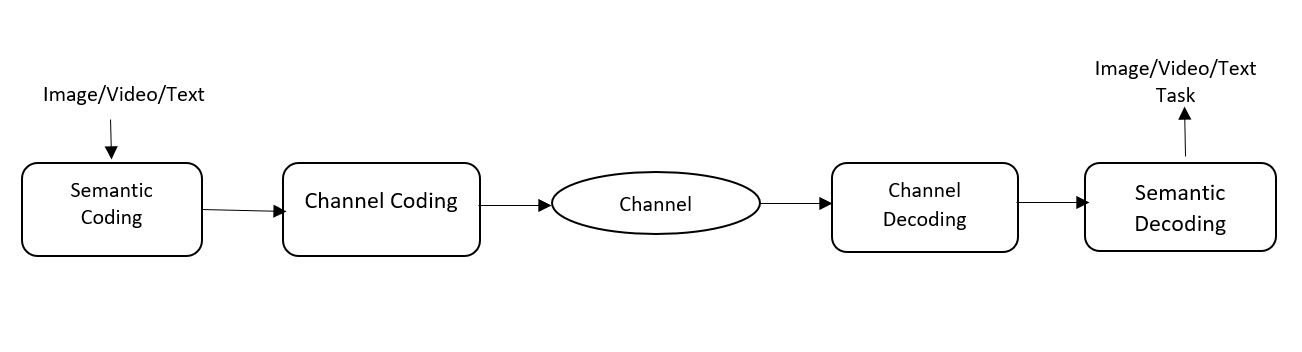}
    \caption{Preliminary semantic communication system}
    \label{fig:System Model}
\end{figure*}

As at now, there is no unique transmission strategy for semantic communication yet, forcing us to design a semantic communication system in accordance with the current communication framework. However, Fig.~\ref{fig:System Model} illustrates a preliminary system model for semantic communication for text and speech transmission. The transmitter extracts the semantic meaning of the message and applies channel coding to the bit stream, then transmits it over the communication channel. On the receiver side, first, the channel decoder decodes the semantic information from the received signal. Then, the semantic decoder takes the semantic information as input and produces output with the existing knowledge base which is shared between the encoder and decoder. Keeping a well-trained knowledge base at the transmitter and receiver is one of the influential factors in a semantic communication system. The challenge is to ensure that the transmitter's semantic information is preserved at the receiver while transmitting through the physical channel. Therefore, a significant amount of research is required for different media types over conventional communication standards with semantic communications.
\par
This research aims to develop a semantic communication system to transmit images over a mobile communication channel to optimize the bandwidth while maintaining the quality of the image. A semantic map is extracted from a given image and it is channel endoded before it is sent over a noisy channel to the decoder. At the decoder the semantic bit stream is channel decoded and use it as an input to the GAN (Generative Adversarial Network) to get the desired image at the decoder. A common knowledge of the images are shared between both the encoder and the decoder. To reduce the amount of data delivered only a semantic map of the image is sent through the channel. As a result, there will be a considerable reduction in demanding energy and wireless bandwidth, resulting in a more sustainable communication network. These features served as inspiration for the construction of a GAN \cite{goodfellow2020generative} based semantic communication system for image transmission.
We evaluate the influence of channel coding on semantic communication in the presence of a noisy channel where polar codes are used as our channel coding scheme. The codewords are modulated using binary phase-shift keying (BPSK)\cite{Omijeh2016BPSK} over an additive white gaussian noise (AWGN) channel. At the receiver end, the noisy log-likelihood ratios (LLRs) are fed to the decoder, and the decoder computes the estimated codewords. We then increased induced noise at the channel to evaluate the effect of white noise on semantic communication and studied three scenarios. In the first scenario, the impact of the quantization noise with zero channel noise is evaluated. The second scenario considers the impact of the varying channel noise on the virtual channel, and finally, the third scenario evaluates the impact of combined quantization noise and varying channel noise levels on the virtual channel. We have derived the maximum amount of channel noise that can be present for effective semantic communication and demonstrate that preserving the edges of semantic maps is crucial in designing future communication systems that rely on semantics. 

The peak-signal-to-noise-ratio (PSNR)\cite{BULL2021335} is calculated for the transmitted images at the receiver side for comparison. Furthermore, JPEG image transmission under similar constraints is considered and results indicated that the proposed system outperform the JPEG compressed system by a significant margin. To evaluate how the semantic features are mapped from transmitter to receiver, we have conducted a subjective experiment using a sample of $30$ users of different socioeconomic backgrounds. This has evaluated the human perceived semantic similarity of the transmitted images in relation to the ground truth images. These experiments fulfill the human and machine perspectives of the developed semantic communication system.
In summary, the main contributions of the paper are summarized as follows: 
\begin{enumerate}
   \item We propose a semantic communication based image transmission system for limited bandwidth wireless communication systems that leverage a  pre-trained GAN network generating the intended image at the receiver that is transmitted as a segmented semantic map of the ground truth image. 
   \item We study the effect of the physical channel noise in designing semantic communication systems with existing physical communication channels and derive important insights. The study also evaluates the effect of channel noise for three scenarios. 
   \begin{enumerate}
	\item Semantic communication of the images under varying channel noise.
 	\item Quantization effect on semantic communication with zero channel noise.
	\item The joint channel noise and quantization effect on semantic image transmission. 
   \end{enumerate}

\end{enumerate}
The rest of the paper is organized as follows. The related work section II gives a brief review of the relevant academic literature in terms of the theoretical aspects, usage of ML techniques for semantic communications, and the usage of GAN in image processing tasks and existing semantic communication based architectures and their limitations. Section III introduces our proposed semantic communication based image transmission system, and it is followed by an analysis of the results in section IV. Finally, in section V we briefly discuss the limitations of the presented work followed by its conclusions in section VI, and future work explained in section VII.

\section{Related Work}

Integration of machine learning(ML) methodologies into 6G communication technology has paved the way for future research in developing end-to-end communication systems based on learning-based optimization about image/video transmission. The 6G wireless communication networks will be the cornerstone of the future of human and machine communication. The exponential development of multimedia content urges the need for wireless communication networks to stand apart from the traditional design paradigm of first to fifth-generation wireless networks of high transmission rates but to have an intelligent link to ML technologies \cite{6gWhitepaper2020}. The paper discusses the application of  ML in different layers of the 6G network, such as the physical layer, medium access layer, and application layer, to provide reliable and time-efficient connectivity to modern applications. It also discusses the security and resource allocation problem in the 6G network.

It is anticipated that semantic communication will become a key paradigm in the development of end-to-end communication systems for 6G networks\cite{semantic20216g},\cite{ZHANG202260},\cite{wang2022transformer},\cite{EdgeSC2022}.Even though semantic communication is predicted to go over the standard Shannon paradigm, there are still several challenges that must be met before modern applications such as the internet of things (IoT) or AR/VR  can be enabled using semantic communications. As a starting point, the goals and the compelling justifications for using semantic communications in 6G \cite{yang2022semantic} are discussed in this research work. An overview of the fundamental 6G ideas and important enabling technologies that underpin semantic communications has been discussed. The paper provides a broad view of the theories of semantic communication and develops semantic communication that is directed into three broad categories in theory, namely semantic-oriented communication, goal-oriented communication, and semantic-aware communication. Most of the initial work is based on semantic communication systems for text or speech transmission\cite{UBT2023}. Authors of \cite{sana2022learning} explore the benefits of semantic compression to go beyond 5G and presents a transformer-based Semantic Communication system. 

Joint transceiver optimization is now achievable with the development of end-to-end (E2E) communication systems with deep learning capabilities that combine all physical layer blocks in conventional communication systems. Deep joint source-channel coding (DeepJSCC) for image transmission is proposed in \cite{deepJSCC2019}, where the image transmission does not depend on the separate source and channel coding, instead the developed convolutional neural network (CNN) directly maps the bits of the image into the channel input symbols. The encoder and decoder are employed by a CNN, which is jointly trained, and the communication channel is a non-trainable AWGN channel. The approach involves transferring a latent representation of the image, rather than extracting semantic information, for the purpose of communication.

\par
With the fast advancement of AI technologies, the focus has been on developing end-to-end image compression systems with deep neural networks  ~\cite{Anelli},\cite{agustsson2019generative},\cite{9065473Patel}. A convolutional neural network combined with an enhanced  JPEG encoder is presented in~\cite{prakash2017semantic}. The model identifies and flags the content of interest, and the visual quality is improved by using a higher bit rate for the content of interest and a lower bit rate for the other regions. In this research, the concept of semantics is being used in terms of image compression. The way forward for image semantics would be to use the semantic properties of images in communication tasks. 

Natural language processing (NLP), powered by deep learning, has had remarkable success in analyzing and comprehending many linguistic documents. DeepSC, a semantic communication system \cite{9398576} built on deep learning for text transmission, was presented in this research. In contrast to bit- or symbol errors in conventional communications, the DeepSC tries to recover the meaning of phrases to maximize system capacity and reduce semantic errors. The crucial aspect is speeding up the joint transceiver training and using the model in different communication settings. To achieve these objectives transfer learning is also employed. DeepSC-S \cite{ST} is a semantic communication system for speech signals that makes use of an attention mechanism and a squeeze-and-excitation (SE) network. The attention mechanism is employed to reduce the distortion of the received signal. The architecture is the designing of the semantic encoder/decoder jointly with the channel encoder/decoder to extract the semantic features, in this context, the important speech signals for the message. Results show that DeepSC-S is more resistant to channel noise, especially in the low signal-to-noise ratio(SNR) regime. DeepSC and DeepSC-S are both semantic communication systems designed for transmitting language-related messages. 

In real-world applications, semantic data is the data that the receiver requires to do its intended task, and often this is not known to the transmitter. To address this problem, a neural network-based semantic communication system~\cite{MLSC1} has been created with a semantic coding network and the data adaptation network (DAN). Semantic coding network learns how to pull out and send information that makes sense by using a method called \say{receiver-leading training}. The DA network learns how to turn the data it has seen into a form similar to the empirical data that the semantic coding network was trained on using transfer learning. As mentioned in the paper, the proposed method does not produce image details as clear as those obtained with the JPEG2000-based method. Additionally, some recovered images exhibit more patches of color contamination at certain SNR levels. The authors concluded that the system is more resilient only in low SNR scenarios.

\subsection{ML for SC}

With the fast advancement of AI technologies, semantic communications systems were enabled with networks, which can learn to extract and transmit the required data depending on the task and channel status. Semantic communication systems for language processing can use three types of neural networks; recurrent neural networks (RNN), CNN, and fully-connected neural networks (FCN).  Xie \emph{et al.}\cite{ 9398576} state that RNNs are lacking in identifying the relationship between the words in long sentences; on the other hand, the  CNNs, because of their small kernel size to achieve computational efficiency, the performance is low. To overcome this, in order to achieve correctness and performance, FCS can be used.

\subsection{GAN}
ML research uses the generative model in various tasks by studying a collection of training samples and learning the probability distribution of the samples and then generating similar data by using the learned probability distribution. GANs may be considered one of the most successful generative models, which were introduced by~\cite{goodfellow2020generative}. Since then, many versions of GANs have been developed for different purposes, such as image-to-image translation, image/video generation, high-resolution image synthesis, classification, and many other computer vision tasks. The rise of GAN-related research since its introduction has resulted in several versions of GAN networks. Authors in \cite{aldausari2022video} have discussed four main architectures of GANs: convolutional GAN, conditional GAN, infoGAN, and AC-GAN. 
\par
As listed in \cite {aldausari2022video}, GANs consist of two networks, a generator, and a discriminator. The term \say{generative} refers to generating new data, and the term \say{adversarial}  refers to the competition between the two networks. In the plain vanilla form, the two networks could be any combination of autoencoders, Fully connected networks, CNNs, and RNNs. The two networks of the GAN are competing in the min-max game. The generator takes a random noise as input and produces new data by learning from the real data, and the discriminator does a classification task by taking the generated data and the real data as inputs. It classifies the real data as \say{$1$} and the generated (fake) data as \say{$0$}. As the name suggests, the discriminator discriminates the fake data samples over the real data, and this competition continues until neither network can improve its performance by changing the learning parameters. 
\par
Conditional adversarial networks have been used as a means to do image-to-image transformation, and an example of such implementation is pix-to-pix \cite{pixTOpix2018}, which is a  conditional adversarial network that generates images based on input label maps, reconstructing objects from edge maps, and colorizing images. Image semantic coding refers to extracting the semantic features of an image. GAN-based photo-realistic high-resolution image generation using semantic maps is discussed in \cite{wang2018high}, and the generated image resolution is as high as 2048 × 1024. A GAN-based image semantic coding system is proposed in \cite{deepImagecoding}, and the performance is compared with conventional image coding techniques of  BPG, WebP, JPEG2000, JPEG, and other deep learning-based image codecs. The previously mentioned GAN-based systems are only used for image generation and are not intended for communication purposes.

\par
The existing body of research suggests that deep learning-based image compression methods are also an extensively researched area in relation to image coding. These systems use autoencoders to transform an image into a latent representation. The framework defined in \cite{agustsson2019generative} uses GAN to develop an image compression system to operate at extremely low bit rates. The system comprises an encoder, decoder/generator, and multi-scale discriminator, and all the components were jointly trained to achieve the goal of learned compression. Semantic generation pyramid \cite{sempyr} is a hierarchical framework that takes advantage of the full spectrum of semantic information encoded in such deep features, from the most basic information in finer features to the most advanced semantic information in deeper features. The approach involves utilizing GAN for image manipulation to realize the objective of compressing images, and not for the purpose of communication. 

\section{Proposed Model}
\par

\begin{figure}[ht]
    \centering
    \includegraphics[width=\linewidth]{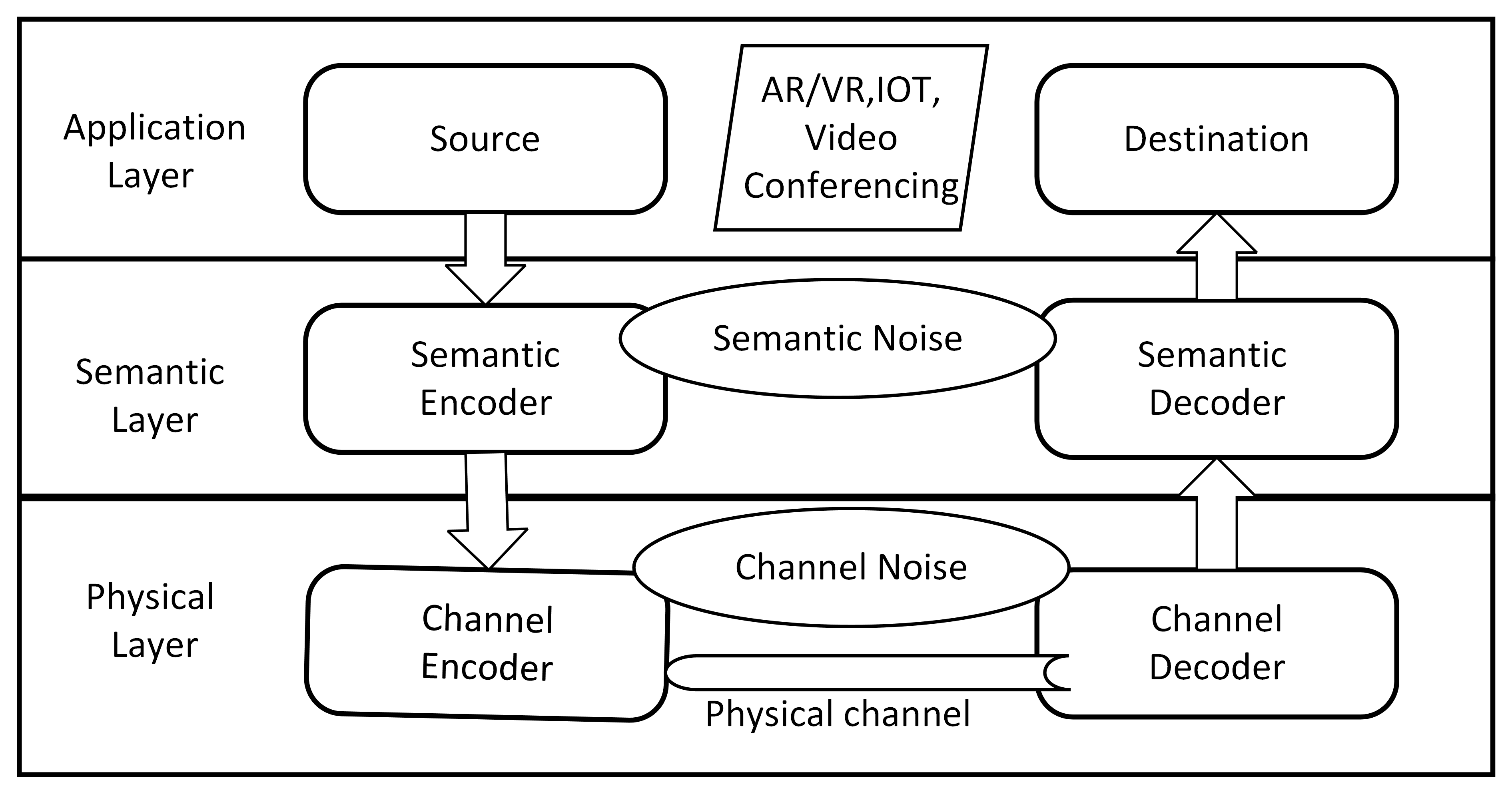}
    \caption{Theoretical three-layered model of semantic communication} 
    \label{fig:Theory}
\end{figure}

The proposed semantic communication-based image transmission system architecture is based on the three-layered theoretical semantic communication model that is derived based on Shanon's theory and is illustrated in Fig. \ref{fig:Theory}. The semantic layer consists of semantic feature encoding and decoding. The semantic encoder does the semantic feature extraction, where the targeted message is text, image, or video. For this extraction process, to interpret the meaning of the message, the encoder utilizes the common knowledge base that is shared between the transmitter and the receiver. The semantic decoder retrieves the meaning of the received semantic symbols using the knowledge base shared between the semantic encoder and decoder. The physical layer, which lies underneath the semantic layer, is concerned with the bit-level transmission of the extracted semantics and employs channel-level optimization to data transmission. The application layer, which lies above the semantic layer, deals with the task-specific details of the received message, such as classification, object detection, scene prediction, etc. The proposed semantic communication system for image transmission is based on the above theoretical model. 

\begin{figure*}
    \centering
    \includegraphics[width=\linewidth]{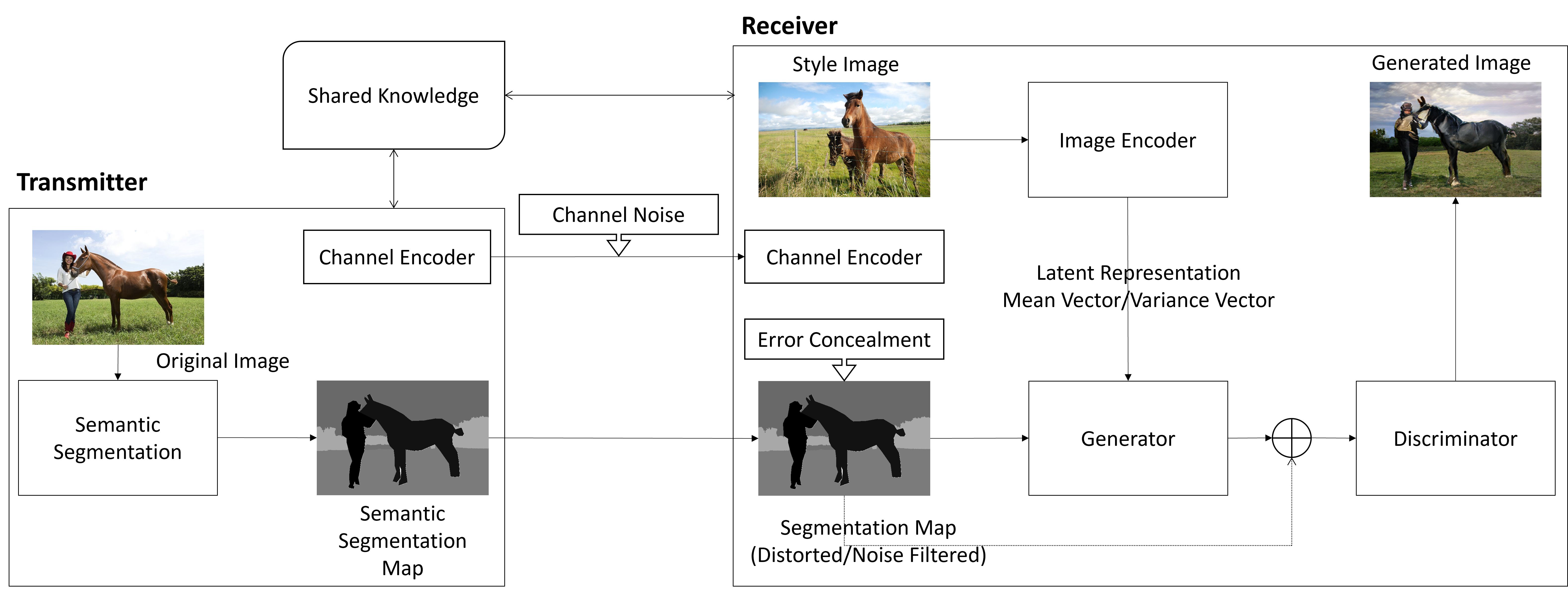}
    \caption{High-level architecture of the proposed image transmission platform using semantic communications}
    \label{fig:HighLevelArch}
\end{figure*}

The high-level architectural diagram of the system is shown in Fig. \ref{fig:HighLevelArch}, and it consists of different blocks which belong to different layers of the aforementioned theoretical framework. The semantic extraction block, common knowledge base, and the GAN belong to the semantic layer. The physical layer comprises of error concealment block, channel encoder, channel decoder, and transmission channel. The error concealment block's usage is to remove any channel noise introduced during the transmission through the physical channel. The channel encoder exerts redundancy on the channel, and the decoder exploits the redundancy to recover the exact bit sequence by facilitating the detection and correction of the bit errors during the physical transmission of the extracted semantic message. 

\subsection{Common Knowledge Base}

 COCO dataset \cite{coco} is the common knowledge base in the context of the developed semantic communication-based image transmission system. The GAN is pre-trained on COCO-Stuff \cite{cocostuff}  dataset, a derivation of the COCO dataset. Common Objects in Context (COCO) is a dataset defined for object detection, segmentation, and captioning. As its name suggests, it is an image repository that comprises day-to-day objects captured by everyday scenes, which consist of $118,000$ training images and 5000 validation images. It contains 182 semantic object classes defined within, and the GAN that we have used in this experiment is pre-trained with respect to these 182 semantic object classes. The segmented semantic maps used in this research are sourced from the above data set, which is theoretically to be semantically extracted in the semantic extraction block. Ten sample images were selected such that it falls into different categories, such as human figures, scenery, animals, vehicles, and household. The intended task at the receiver, which belongs to the application layer, is a classification task in which the receiver classifies the objects according to the objects on which the GAN was trained.       

\subsection{Semantic Decoder}

\begin{figure*}[ht]
    \centering
    \includegraphics[width=\linewidth]{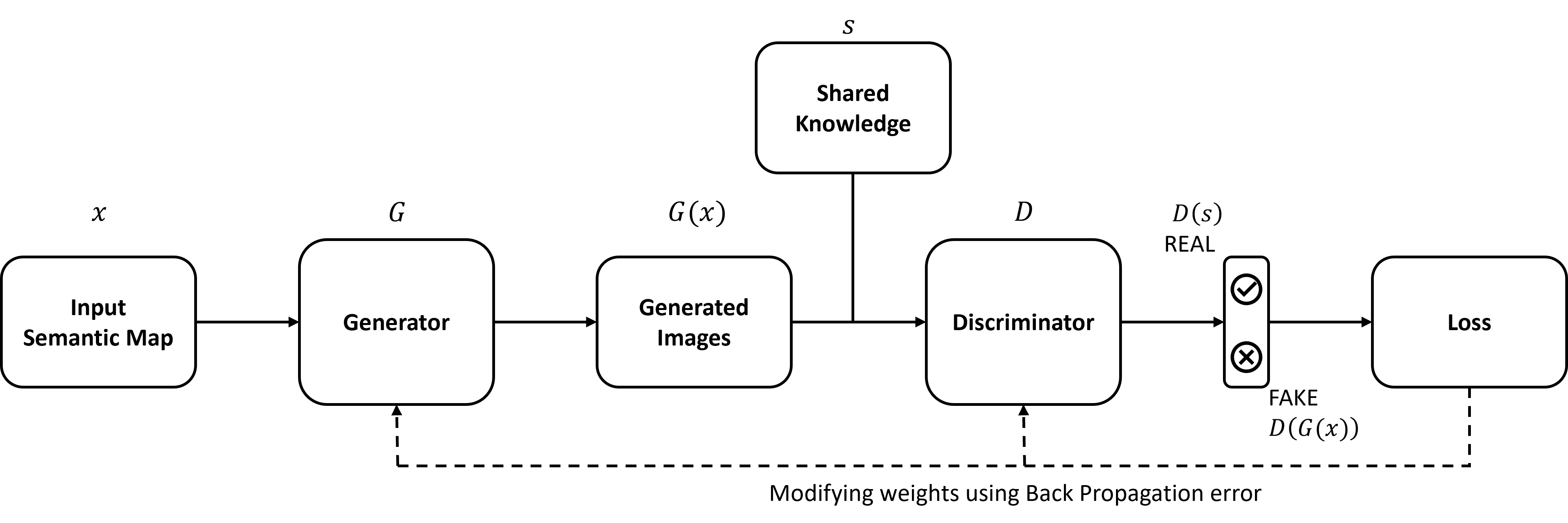}
    \caption{GAN Architecture}
    \label{fig:GANArch}
\end{figure*}

GAN used in this research is based on semantic image synthesis with specially adaptive normalization \cite{Semantic},which is called the SPADE network. From Fig. \ref{fig:GANArch}, the generator $G$ produces the images $G(x)$ from a random input $x$. These produced images are then fed into the discriminator along with the real images from the training data. The images are then classified as real or fake by the discriminator model. The weights of the generator and discriminator must then be updated when the loss is calculated, and this loss needs to be back-propagated. As a result, each epoch sees the generator and the discriminator improving.

The term "$p_{data}(s)$" is the probability distribution of real images, and "$p_x(x)$" is the probability distribution of fake images. $E_{s\sim p_{data}(s)}[log{D(s)}]$ is the average log probability of the discriminator when the real image from the shared knowledge is input and $E_{x\sim p_x(x)}[log{(1-D(G(x)))}]$ is the average log probability of the discriminator when the generated image is input. The discriminator aims to enhance the V(D,G), whereas the generator aims to reduce the V(D,G). The discriminator aims to accurately identify the real and fake images in order to optimize the loss function V(D,G)\cite{aldausari2022video} as calculated in equation \ref{eq:1}.

\begin{equation} \label{eq:1}
\resizebox{.91\hsize}{!}{$\min\limits_{G}\max\limits_{D}{V(D,G)}=E_{s\sim p_{data}(s)}[log{D(s)}]+E_{x\sim p_x(x)}[log{(1-D(G(x)))}]$}
\end{equation}

The discriminator seeks to accurately categorize real images as true in order to bring D(s) as near to 1 as feasible by maximizing the term $E_{s\sim p_{data} (s)} [logD(s)]$ and seeks to maximize $E_{x\sim p_x (x)} [log⁡(1-D(G(x)))]$ by accurately categorizing false pictures as such and bringing $D(G(x))$ as close to 0 as feasible. The generator tries to trick the discriminator by producing images that resemble real images in an effort to reduce the loss function V(D,G). The generator aims to minimize the term $E_{x\sim p_x (x)}[log⁡(1-D(G(x)))]$ by making $D(G(x))$ as near to 1 as feasible.

Vanilla GAN cannot control the images being generated, and hence we have used conditional GAN in our architecture, where it adds a condition to the GAN, and this condition is the semantic segmentation map of the ground truth image. The image generated is matched to the condition provided. The loss function is modified \cite{CGAN} as shown in equation \ref{eq:2}, where c is the condition: 

\begin{equation} \label{eq:2}
\resizebox{.91\hsize}{!}{$\min\limits_{G}\max\limits_{D}{V(D,G)}=E_{s\sim p_{data}(s)}[log{D(s|c)}]+E_{x\sim p_x(x)}[log{(1-D(G(xc)))}]$}
\end{equation}

\subsection{Communication Framework}
\begin{figure*}
    \centering
    \includegraphics[width=\linewidth]{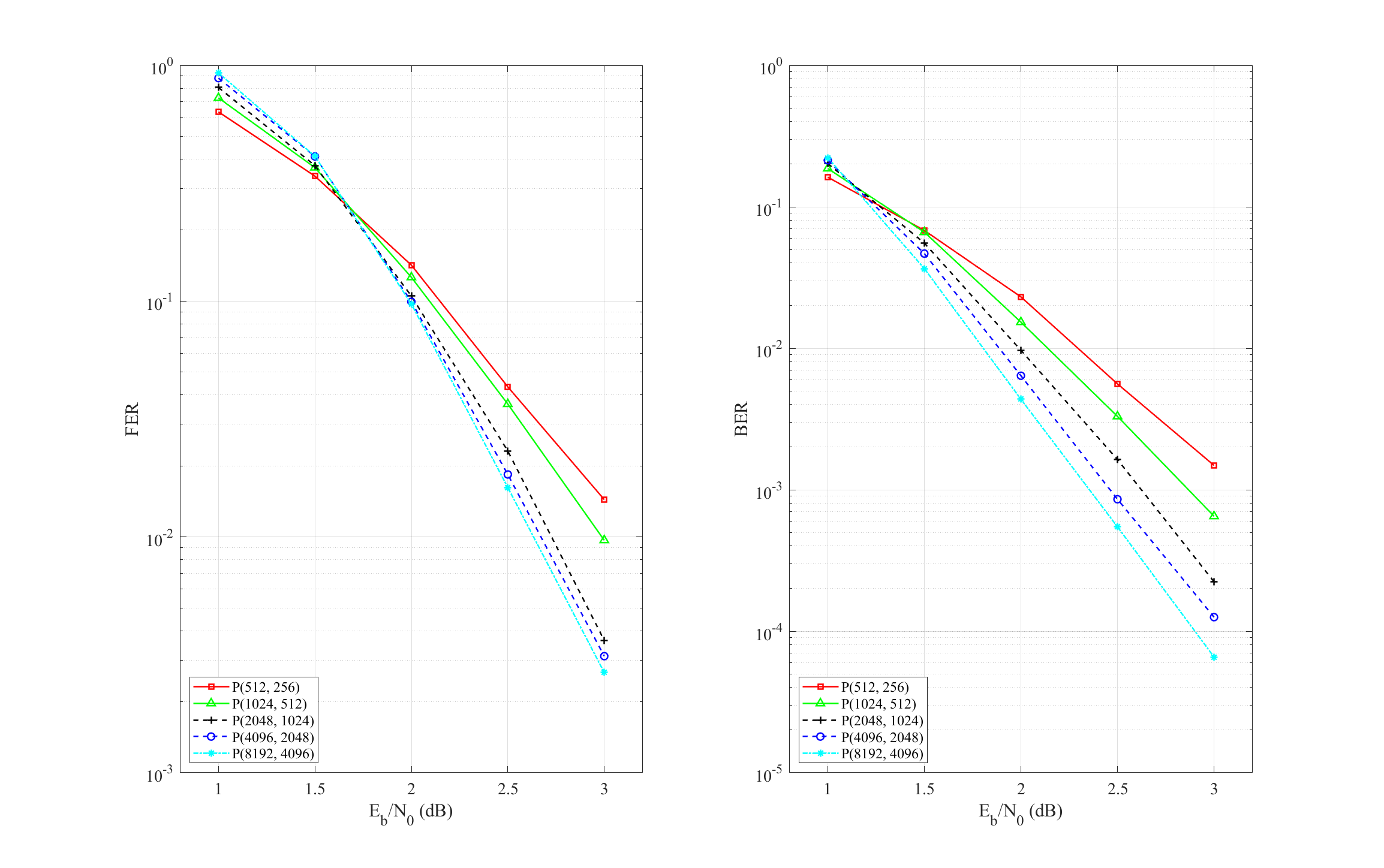}
    \caption{The error correction performance of polar codes with different block lengths.}
    \label{fig:ECCfig}
\end{figure*}

Polar codes \cite{Arikan, Rezaei2022, Rezaei20222, Rezaei20223, Combinational2023} are represented as $\mathcal{PC}(N,K)$ where $N$ and $K$ stand for the block length and the number of message bits, respectively. Channel polarization \cite{Arikan} is proposed as a method to transform the physical channel into extremely reliable and extremely unreliable virtual channels as the code length approaches infinity. In other words, assuming the symmetric channel capacity of a binary-input discrete memoryless channel (B-DMC) $W$ as $I(W)$, the reliability of each individual channel $W_{i}^{N}$ ($1 \leq i \leq N$) approaches to either one (highly reliable ($I$($W_{i}^{N}$) $\rightarrow 1$)) or zero (highly unreliable ($I$($W_{i}^{N}$) $\rightarrow 0$)). The reliable channels can be computed using Bhattacharya parameters \cite{Arikan} where the information bits are located. Fig. \ref{fig:ECCfig} depicts the error-correction performance of polar codes for blocklengths in the range of 512 to 8192 with rate $\mathcal{R} = 1/2$. Obviously, the performance of polar codes improves as the code length grows.

The specification of the selected channel (de)coding scheme is detailed in Table \ref{tab:chnlSpec}. We consider polar codes as our channel coding scheme. To verify the proposed semantic communication system using practical codes, a polar code of size $N=4096$ carrying $K = 2048$ information bits is selected. Similar to \cite{Ercan2017}, the chosen polar code is optimized for $E_b/N_o = 2.5$ dB. The codewords are modulated using BPSK over an AWGN channel. At the receiver end, the noisy LLRs are fed to the decoder, which computes the estimated codewords. 

\begin{table}
\centering
\caption{Channel (de)coding specifications}
\begin{tabular}{ll}
\hline
\textbf{Parameter} & \textbf{Value} \\\hline
Channel en/decoder& Polar Codes\\
\# of information bits&2048  \\
\# of codeword bits& 4096\\
Code Rate& $1/2$\\
Modulation&BPSK \\
\# of bits per symbol & 2\\
Demaping method& Log-liklihood ratios\\
Channel & AWGN\\
\hline
\end{tabular}
\label{tab:chnlSpec}
\end{table}
\subsubsection{Channel and Quantization Noise}
\begin{figure}
    \centering
    \includegraphics[width=0.9\linewidth]{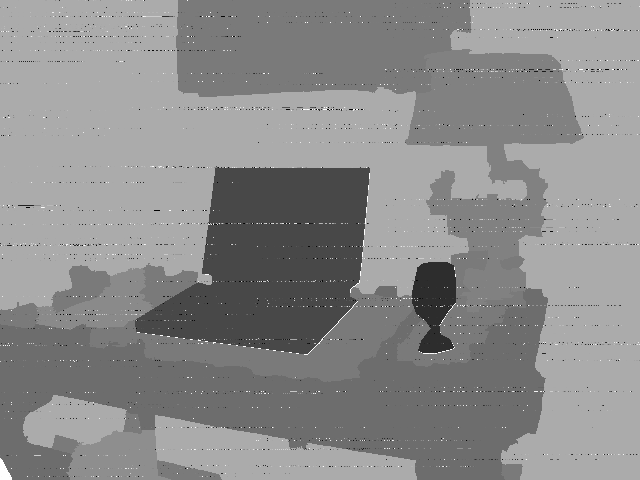}
    \caption{Effect of channel noise on the semantic segmentation map image}
    \label{fig:rcvdimg}
\end{figure}

The information data that needs to be transferred through the channel consists of semantic segmentation map images, as depicted in Fig.~\ref{fig:HighLevelArch}. Each pixel is represented by 8 bits. Therefore, $K/8 = 256$ pixel values can be embedded into a packet. The set of pixels is selected row-wise, and decoding follows. The decoder estimates the codewords at the receiver end, and a software program regenerates the images. Fig. \ref{fig:rcvdimg} presents a semantic segmentation map image regenerated after transferring through the channel for $E_b/N_o = 2$ dB. Obviously, there are bursts of errors in a few image rows. This stems from the fact that polar codes are recursive in nature. Therefore, noisy bits affect the entire packet. 

\section{Results and Discussion}
The following section demonstrates the experimental results gained with the proposed semantic communication-based image transmission system. We have used several scenarios to understand the effect of the different conditions of the physical channel and the transmission formats on the semantic communication model. In the first scenario, we tested the semantic communication of ten images under varying channel noise, the second being the quantization effect on the semantic communication with zero channel noise. The third scenario is the joint channel noise and quantization effect on semantic image transmission.

We have derived insights into the above mentioned scenarios by measuring the PSNR for ten sample images for different channel noise levels. We evaluated the performance of the system by comparing it to the JPEG standard. This was done by encoding the ground truth image using the JPEG encoder, adding channel noise to the encoded bit stream, and then decoding the bit stream. Furthermore, it is important to understand the human perception of the transmitted images in relation to the original images. Hence, to fulfill this objective, a subjective experiment is conducted.

\begin{figure}
    \centering
    \includegraphics[width=0.9\linewidth]{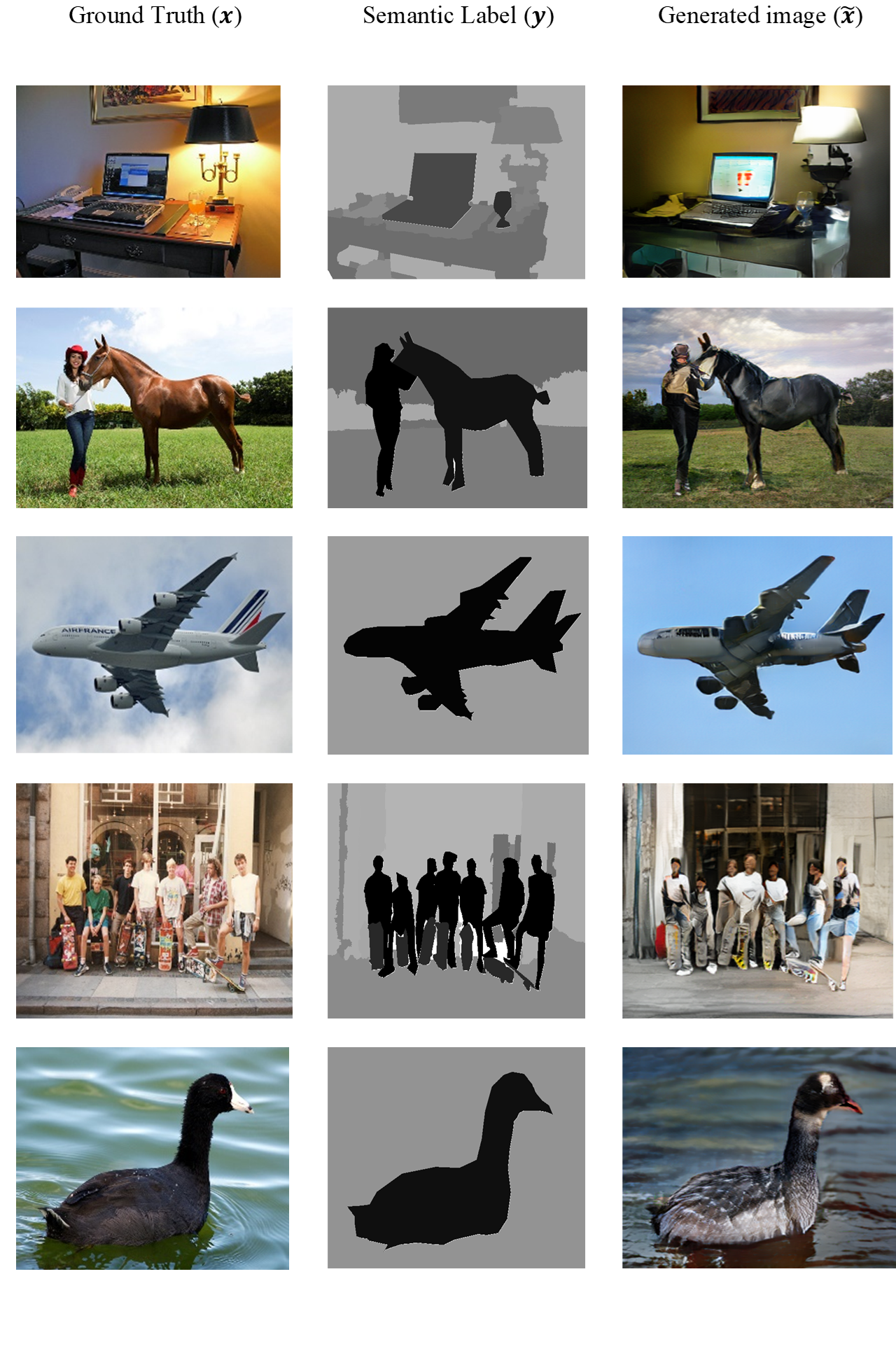}
    \caption{Illustration of the transmitted images}
    \label{fig:images}
\end{figure}

\subsection{Demonstration of the concept}
\par
The notable difference between our communication model with the existing communication model architecture is the extraction of the semantic meaning of the message. Object mapping is considered the semantic meaning of the image. In images, semantic segmentation is associating a label or a category with every different pixel with the help of a deep learning algorithm. It is used to identify a collection of pixels that makes an object recognizable to a specific object class. With the developed system, we have used semantic segmentation maps available with the COCO dataset for our experiment. Fig. \ref{fig:images} illustrates some of the ground truth images (${x}$) and the semantic segmentation map (${y}$) used for the testing setup. The image intended to be transmitted  (${x}$) is semantically segmented (${y}$) and transmitted through the virtual channel. As can be seen from the comparison between the ground truth image and the transmitted image ($\tilde{x}$) to the transmitter, the object mapping is done as intended, although some features, like the light beam, are not exactly reconstructed at the receiver that can be seen from the first image with the lamp. This kind of feature may not be necessary for M2M task-oriented communications when the communication system is trained toward a specific task. 

\subsubsection{Comparison with JPEG image Compression}

\begin{figure}
    \centering
    \includegraphics[width=\linewidth]{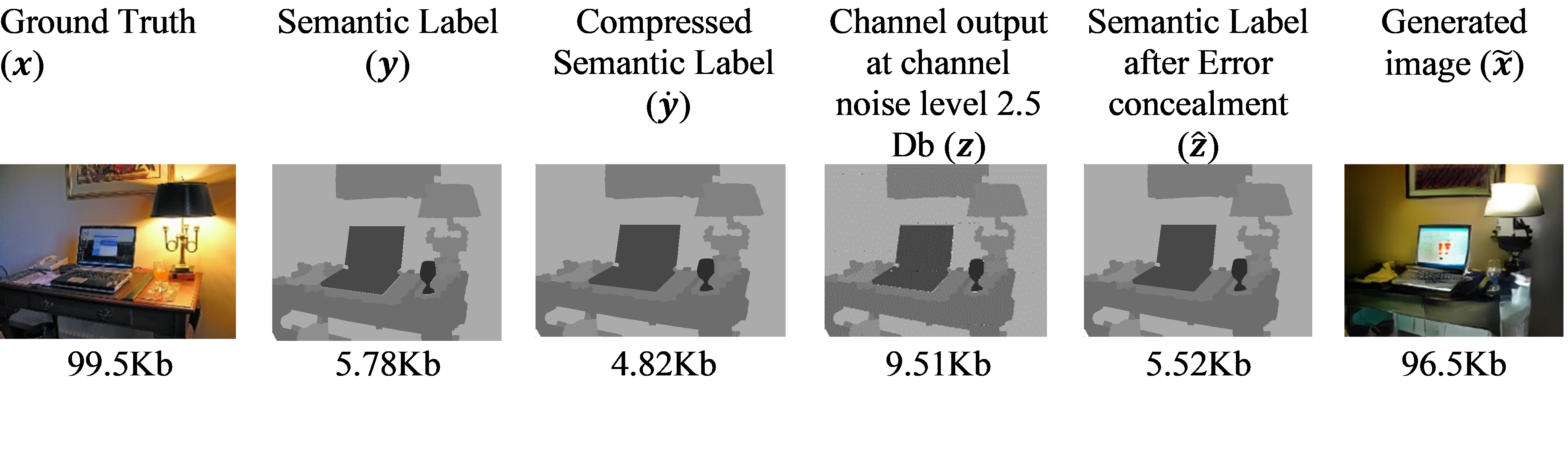}
    \caption{Transmitted image size comparison}
    \label{fig:compression}
\end{figure}

To compare traditional image compression and the proposed communication system, we have considered the JPEG image and the semantic image at different stages of the communication process. The typical compression ratio of JPEG \cite{panchanathan1996jpeg} is around $1:10$ with a small quality degradation. Achieving higher compression costs the quality loss of the image. Fig. \ref{fig:compression} illustrates the image size comparison at different stages of the communication path for the proposed transmission system. The JPEG image intended to be transmitted $99.5$ Kb in size, and the semantic segmentation image was only $5.7$ Kb in size, which results in a compression ratio of 17.2 (${x}$ /${y}$). To further reduce the data rate, image compression is applied on top of the semantic segmentation map to get a further compression gain of $20.6$ (${x}$ /$\hat{y}$). The segmentation map size at channel output is $9.51$ Kb at a noise level of $2.5$ dB. The segmentation map size after error concealment is $5.52$ Kb at this noise level, while the final output image has a size of $96.5$ Kb.

\begin{figure}
    \centering
    \includegraphics[width=\linewidth]{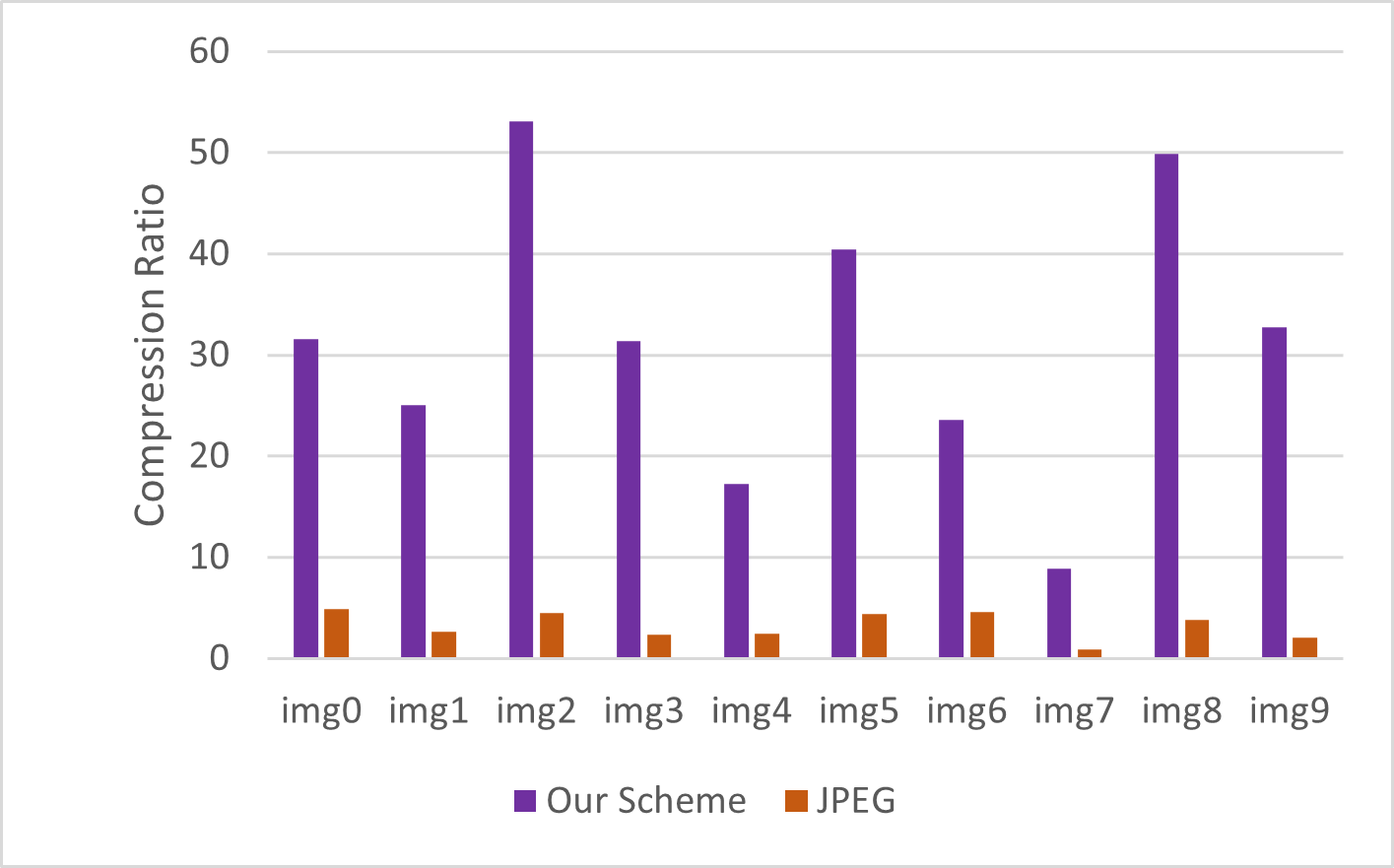}
    \caption{Compression ratio for sample images}
    \label{fig:JPEGcompression}
\end{figure}

\begin{figure}
    \centering
    \includegraphics[width=\linewidth]{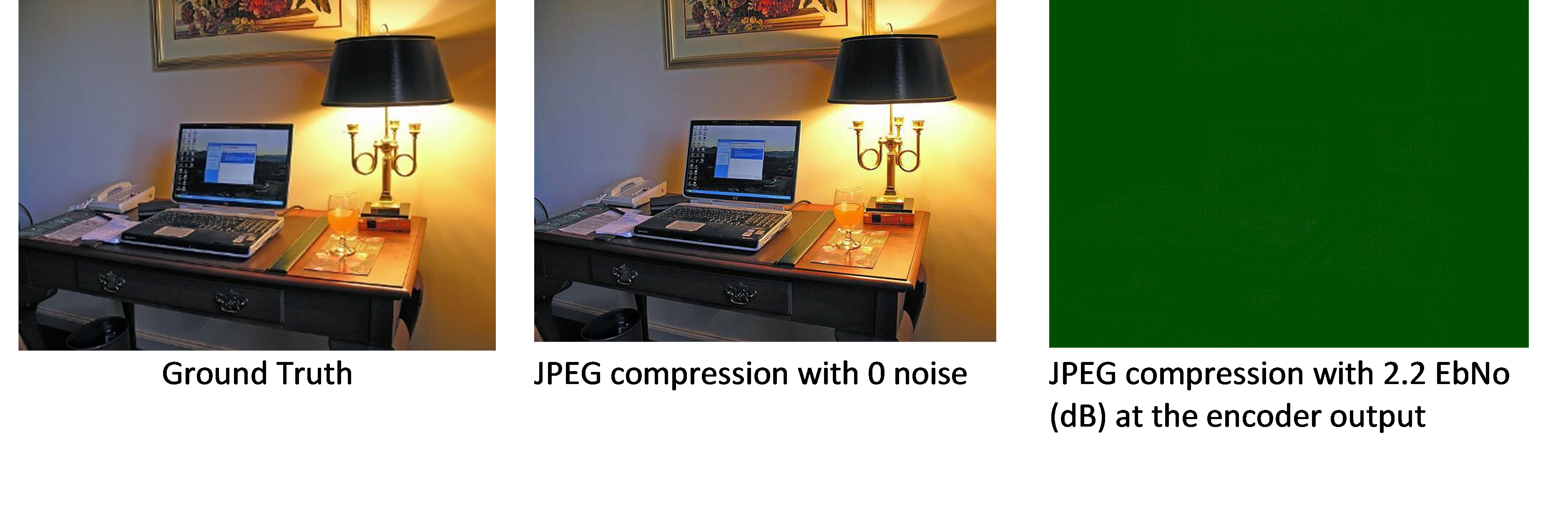}
    \caption{Comparison of JPEG compression}
    \label{fig:JPEGimg}
\end{figure}

The ground truth image was compressed with JPEG compression with zero channel noise scenario and we have plotted the compression ratio for the selected sample images against our scheme and Fig.\ref{fig:JPEGcompression} illustrates the results. Noise is added to the compressed JPEG bit stream and tried to decode the compressed bit stream at the receiver. However, the JPEG decoder fails to decode any bit stream with added noise. Fig.\ref{fig:JPEGimg} illustrates the comparison of JPEG compression with noise. JPEG compression performs well in high Signal-to-noise ratio (SNR) scenarios at the expense of a higher bit rate. The purposed scheme outperforms JPEG compressed image transmission under low bit rate and low SNR scenarios. Even at high SNR scenarios, the proposed compression beats JPEG compression in terms of the bit rates though JPEG compressed images provide better quality for human perception. Since the intended main application of the proposed scheme is to consider M2M communications, the proposed scheme outperforms JPEG compressed transmission for all SNR values since it only consumes 5\% of the bandwidth JPEG used.

\subsubsection{Edge Preservation of Semantic Maps}
\begin{figure*}
    \centering
    \includegraphics[width=\linewidth]{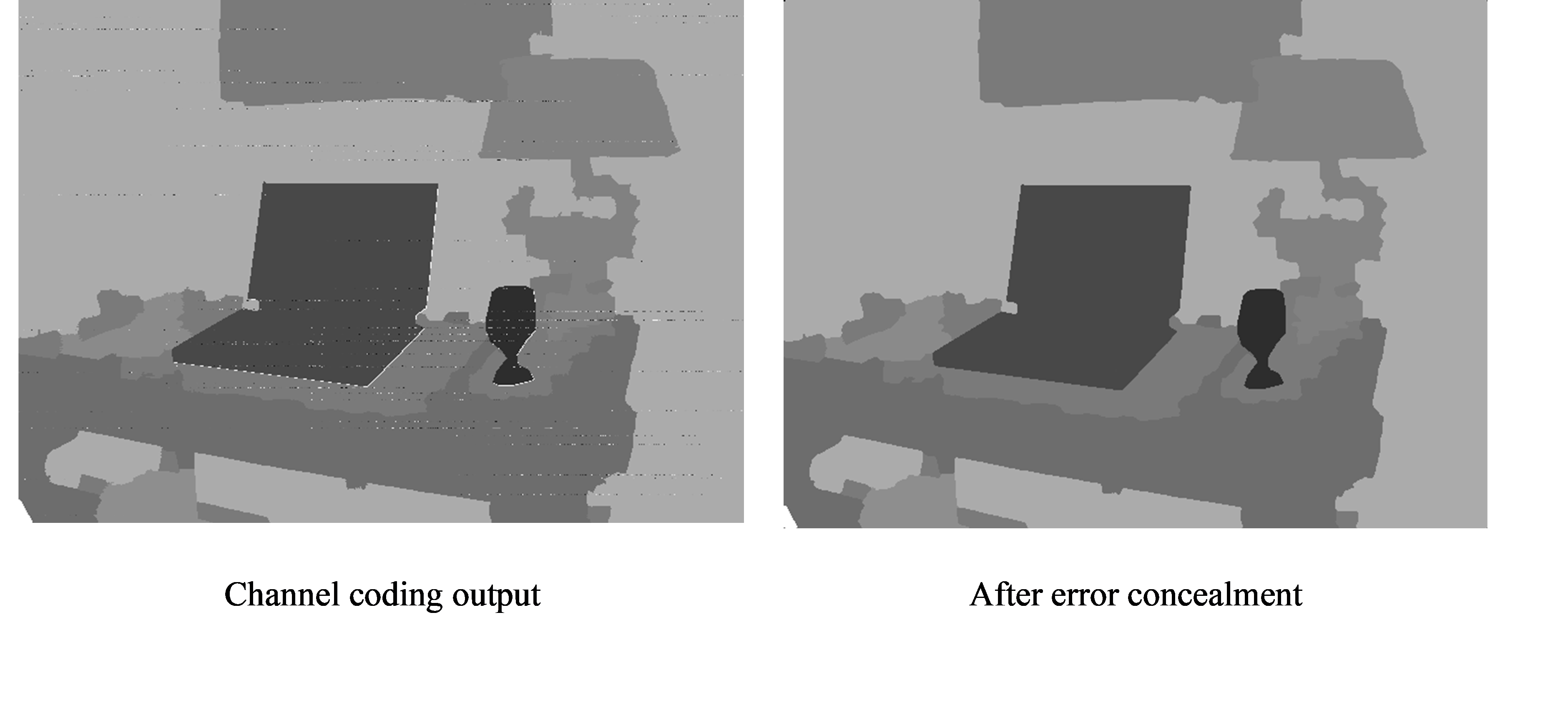}
    \caption{Effect of median filter on salt and pepper noise and edge preservation}
    \label{fig:MedianFilter}
\end{figure*}

In semantic segmentation, each object is identified by a different pixel value, and the difference in pixel values identifies the edges. Object edge distortion during the transmission through the physical channel may cause problems reconstructing the intended image at the receiver. This concept is demonstrated by increasing the physical channel noise from $2.0$ dB to $3.0$ dB, where most semantic segmentation map images failed at the GAN due to severe distortions to the segmentation map. The reason being GAN interprets the shape of the objects differently than the original object based on the identified pixel values and tries to relatively generate the image based on the predefined label classes on which the GAN was trained. Then if the Edges of the objects in the semantic map are distorted beyond a certain threshold, the image interprets more label classes than the number of label classes that the GAN was trained based on the label classes defined in the dataset. With this set of experiments, we have derived that the preservation of the edges is of great importance in the semantic communication of image data. Fig. \ref{fig:MedianFilter} illustrates the channel output at $2.3$ dB channel noise level before and after the salt and pepper noise filtering, which depicts the edge distortion and edge preservation after the median filtering. At the receiver side, the error concealment block is set up to filter out the salt and pepper noise of the received semantic map. We have used median filtering \cite{medianFilter}, a non-linear operation that is very effective in filtering out salt and pepper noise and preserving the edges.

\subsection{Impact of Quantization Noise}

\begin{figure}
    \centering
    \includegraphics[width=\linewidth]{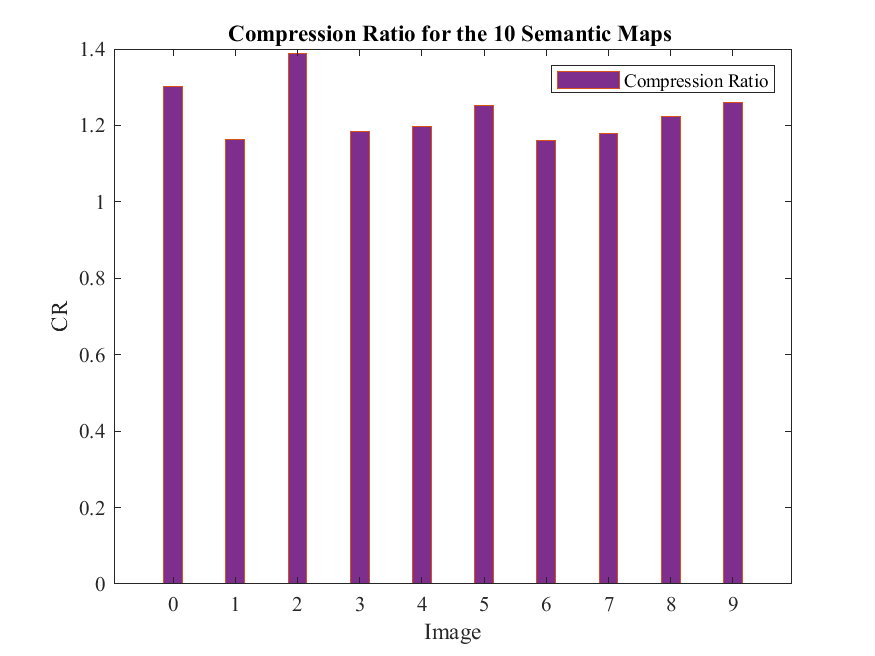}
    \caption{The Compression Ratio of the Semantic Segmentation Map Images } 
    \label{fig:CR}
\end{figure}

We have employed a lossless compression of the semantic maps to understand the effect of the quantization noise on semantic communication. The compressed maps were transferred under zero noise in the physical channel, and the effect was measured. The compression ratio and the PSNR are calculated. Fig. \ref{fig:CR} illustrates the compression ratio for the semantic segmentation map images. The PSNR for all the compressed semantic maps converged to infinity.GAN performed as expected with the lossless compression for the semantic maps considered. To make a fair comparison, we have tried to do the same with JPEG-compressed semantic segmentation map images. However, GAN fails to generate the images since the label classes fall more than $182$ label classes. Therefore it can be concluded that with the lossy compression, the semantic maps' distortion cannot be handled by the GAN, where it was trained based on the $182$ semantic label classes which are defined in the COCO dataset. We may derive that, for a highly versatile semantic communication system, the knowledge base on which the GAN is trained and shared between the receiver and transmitter shall be a highly diversified dataset with different label classes. This result can be more optimized if the GAN's trained dataset is task oriented.

\subsection{Impact of Channel Noise under varying conditions}

\par
\begin{figure*}
    \centering
    \includegraphics[width=\linewidth]{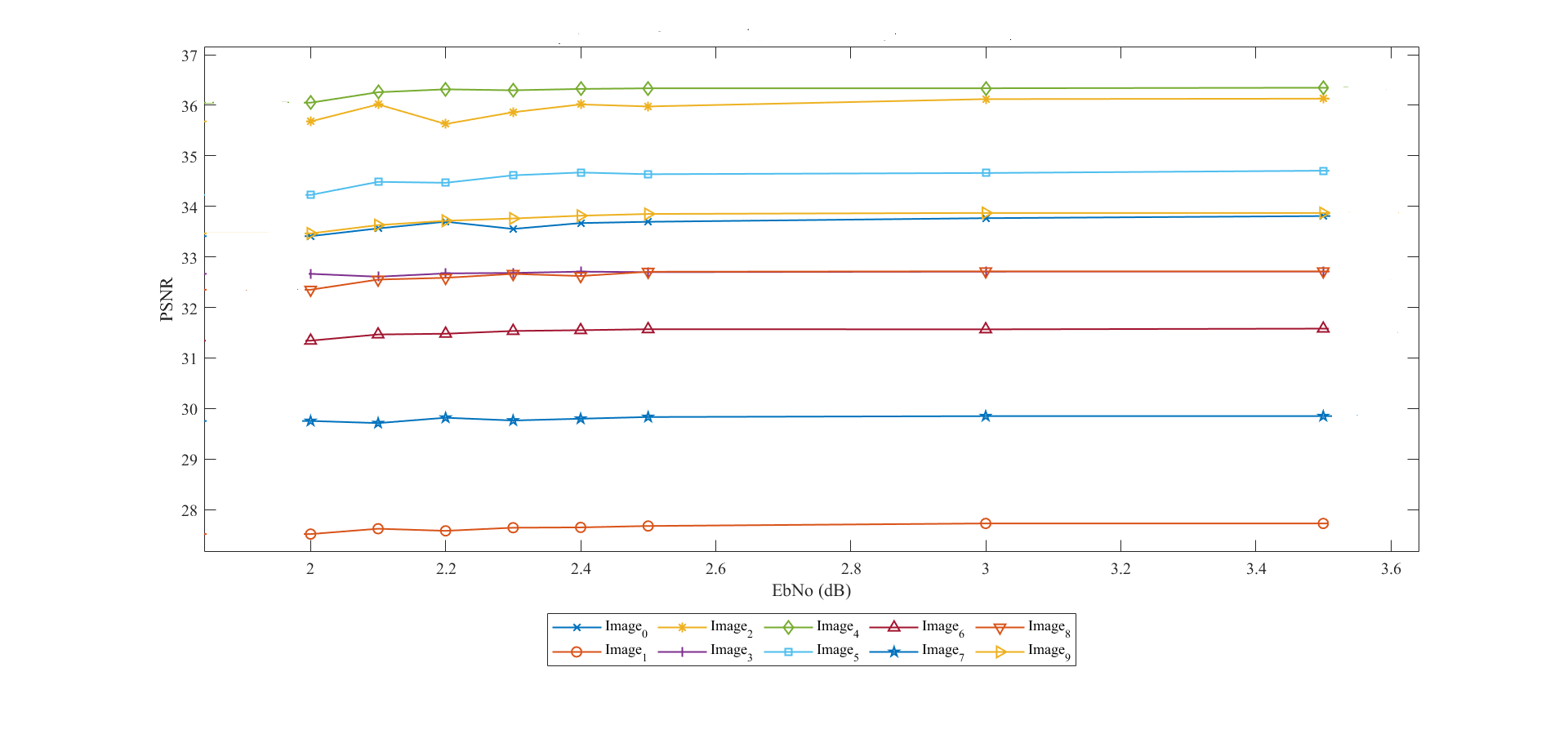}
    \caption{Image quality after noise concealment at different noise levels for sample semantic segmentation map  images}
    \label{fig:PsnrLabel}
\end{figure*}
\begin{figure*}
    \includegraphics[width=\linewidth]{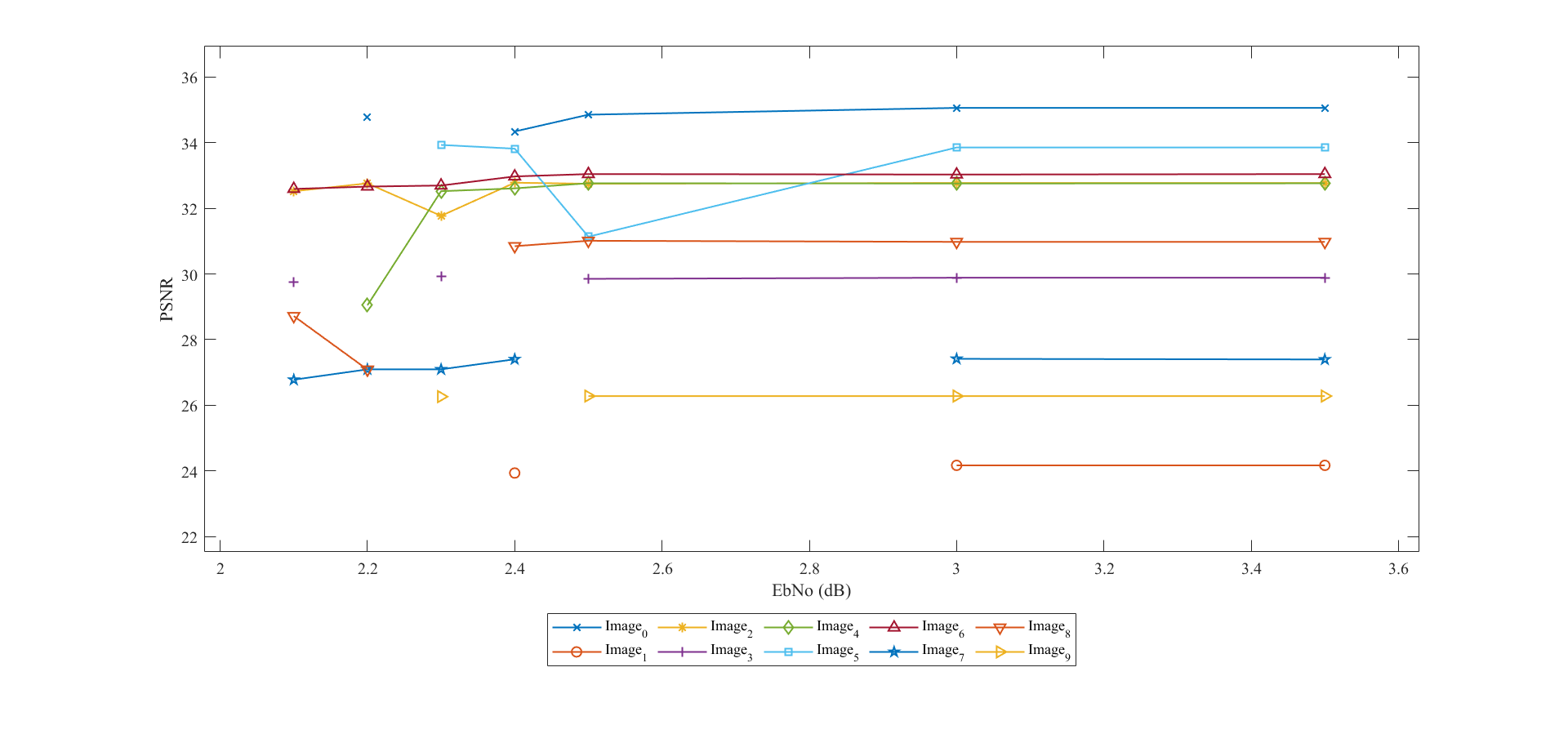}
    \caption{Image quality at different 
 noise levels for generated images} 
    \label{fig:PSNRgenerated}
\end{figure*}

The experimentation on the proposed system is based on the AWGN channel, where the channel adds white Gaussian noise to the signal that passes through it. The added noise has considerably distorted the semantic segmentation map as the GAN failed with the input map. To overcome this issue, median filtering is applied to the received image.  
 It is observed that the PSNR value for a particular segmentation map is the same, with all the channel noise levels from $2.1$ dB to $3.0$ dB after passing through the noise concealment block as illustrated in Fig. \ref{fig:PsnrLabel}. The reference image considered for this PSNR calculation is the semantic segmentation map image with zero channel noise level. When the channel noise level increases, some semantic segmentation map images fail to generate the image at the GAN output. However, the added noise is filtered by median filtering because of the severe object edge distortion. These failed cases are random, and the GAN needed to be trained on a much larger dataset. Although some over-cases failed at the GAN level, this designed system can withstand up to $2.1$ dB noise to generate realistic images at the GAN level. Hence, the system is capable of doing semantic transmission. Fig. \ref{fig:PSNRgenerated} illustrates the PSNR variation for the generated images. 

\par
\begin{figure*}
    \centering
    \includegraphics[width=\linewidth]{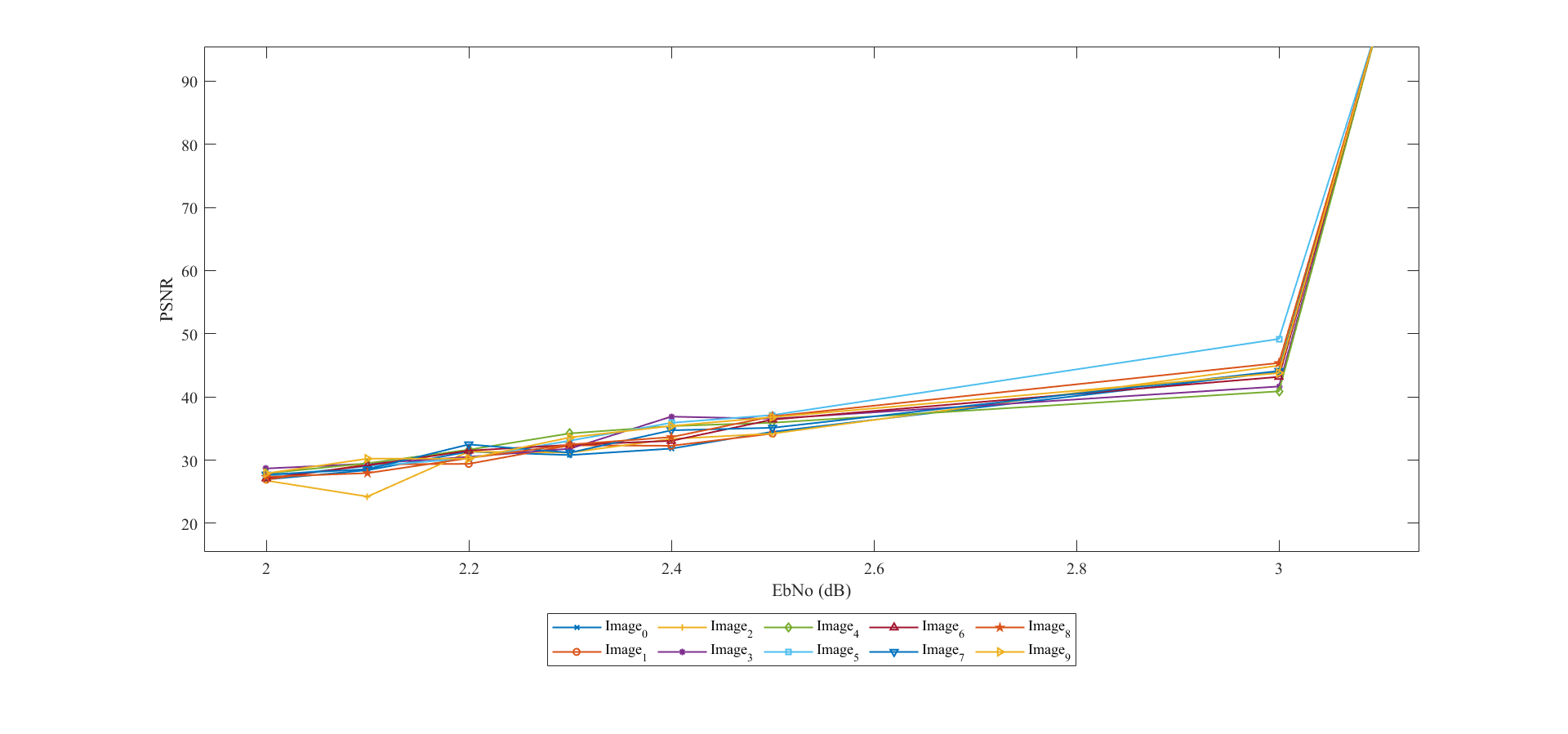}
    \caption{Image quality at different noise levels without noise concealment for semantic segmentation map images}
    \label{fig:PsnrNoFilter}
\end{figure*}

We have transmitted the semantic segmentation map after removing the error concealment block at the receiver and obtained the results. However, GAN fails at all the channel noise levels except the perfect channel and fails to generate images based on the semantic maps since the edge distortion of the objects is severe. Fig. \ref{fig:PsnrNoFilter} illustrates the PSNR values for the selected $10$ semantic segmentation map images without error concealment(salt and pepper noise filtering). Obviously, PSNR decreases as the noise at the receiver increases.

\subsection{ Impact of Joint Quantization and Channel Noise}

\begin{figure*}
    \centering
    \includegraphics[width=\linewidth]{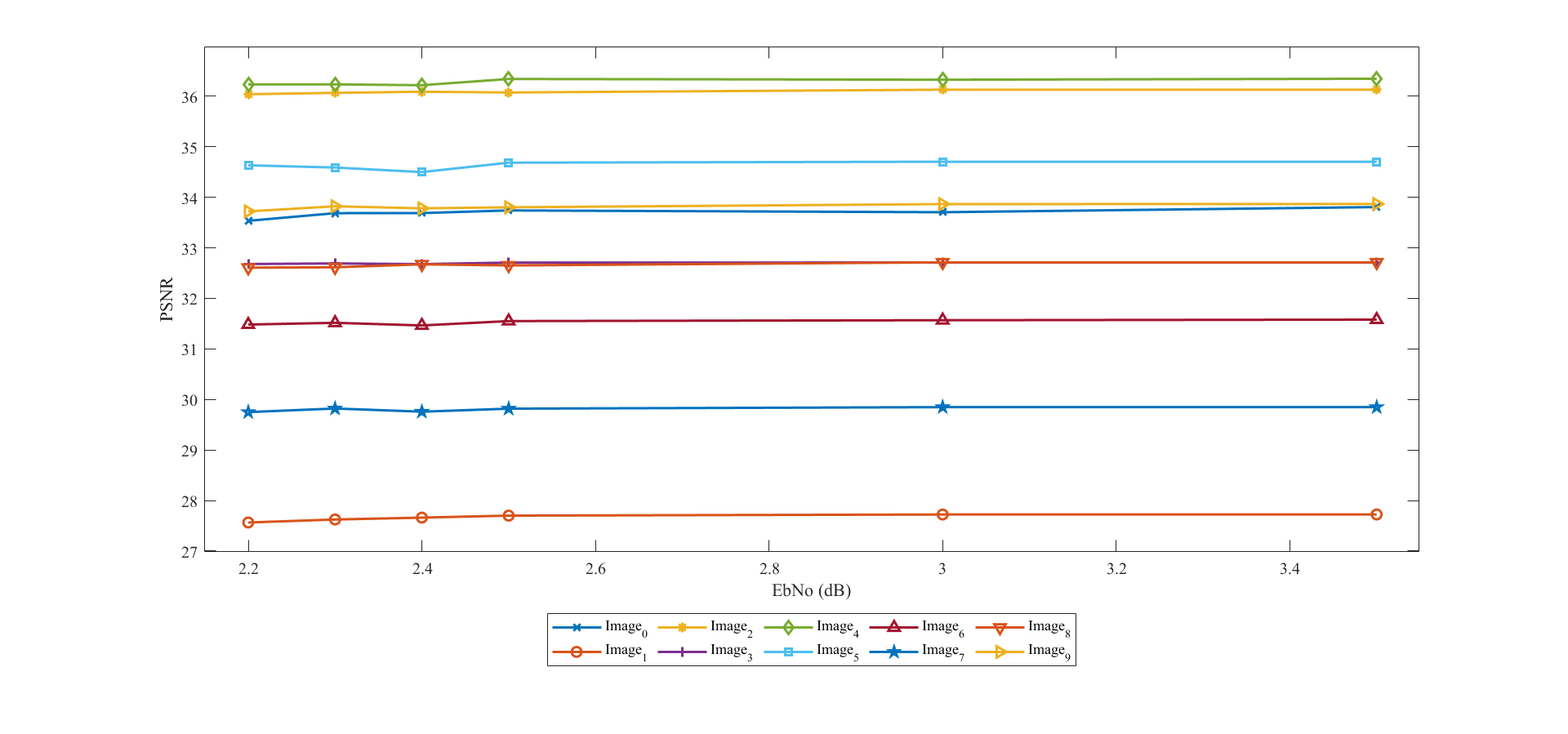}
    \caption{Impact of combined quantization and channel noise for the quality of semantic segmentation map images with noise concealment}
    \label{fig:PsnrCompressLabel}
\end{figure*}

\begin{figure*}
    \includegraphics[width=\linewidth]{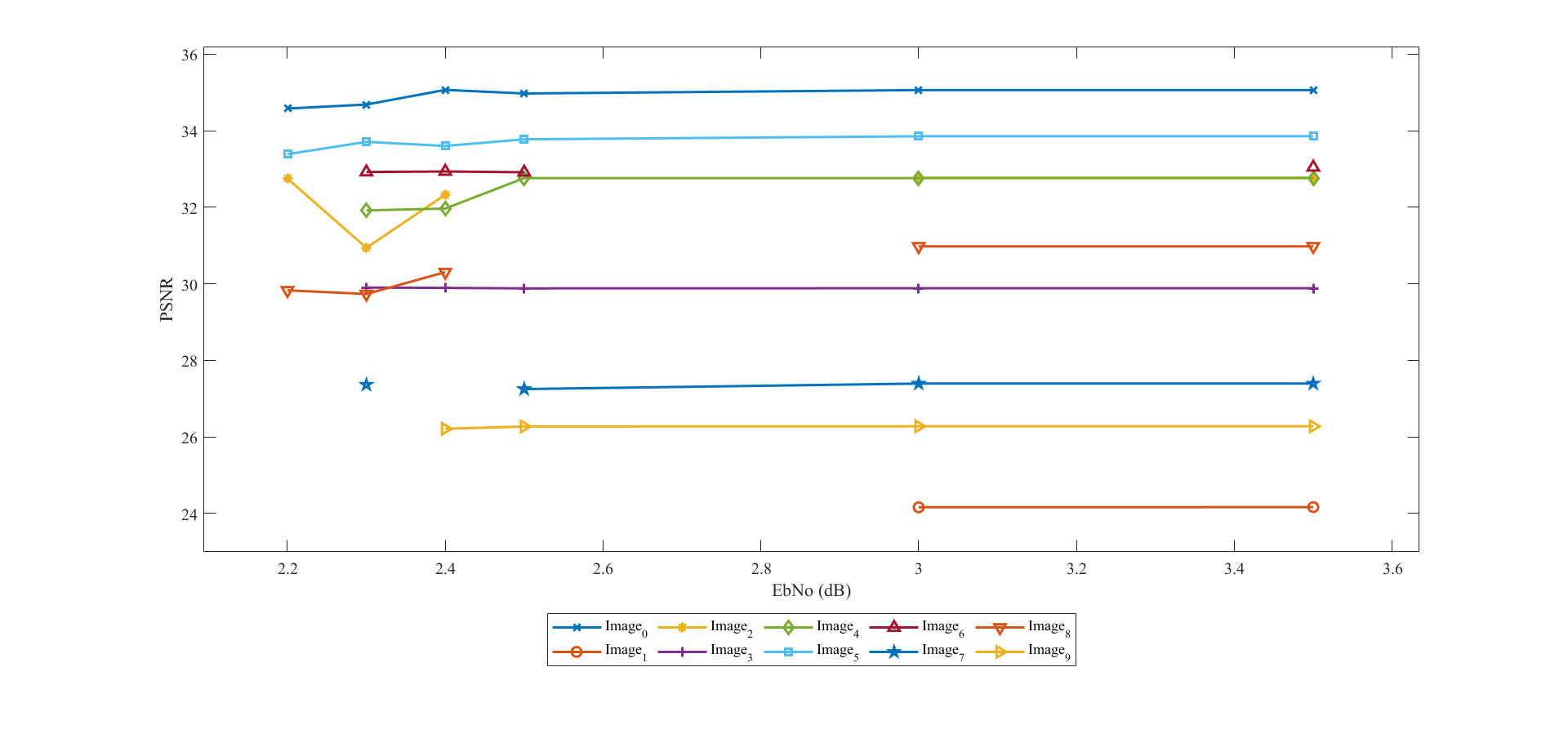}
    \caption{Image quality at different noise levels for generated images with joint Quantization and channel noise} 
    \label{fig:PSNRCompressgenerated}
\end{figure*}

In this scenario, we have studied the impact of joint quantization and channel noise. The transmitted semantic image is compressed at the transmitter side with the lossless compression method. The PSNR values computed for the $10$ images are the same for all the noise levels for a particular semantic label image, as illustrated in Fig. \ref{fig:PsnrCompressLabel}. Although the PSNR of the images being the same after the error concealment with the median filter, the GAN performs differently when the noise level of the physical channel is increased. The semantic communication system described in this paper can withstand up to $2.2$ dB noise level to generate realistic images at the GAN. Some semantic segmentation map images in this range fail to generate an image, where the GAN fails with the error of more label classes than defined. The reason for this is concluded as the edge distortion, which cannot be concealed by the median filter. Fig. \ref{fig:PSNRCompressgenerated} depicts the PSNR for the GAN-generated images. The GAN fails to generate the image for some noise levels because of the edge distortion.   

\subsection{ Qualitative Evaluation}
We use subjective experiments to evaluate the perceived semantic fidelity \cite{shi2021new} by human users for the generated images by the GAN, where the images technically are transmitted images with the Semantic communication system. The test was conducted to compare the semantically transmitted image with the ground truth image for its semantic similarity. We used a sample of $30$ users for the experiment, and the sample was selected randomly, representing various educational and social backgrounds. Users were given instructions to consider the semantic information of the ground truth image and the generated image by the GAN and asked to rate the generated image on a scale from $1$ to $5$,  where $1$ represents the least semantic fidelity and $5$ represents the highest semantic fidelity. The users have explained the task that they would do. They were given the following explanation on their questionnaire:  
\say{Image B is a Computer-Generated Image based on the features of Image A. Rate the resemblance between the two images. For example, in comparison $1$, do all the objects in the original image are represented in the correct position in the generated image, and the sizes and features of the objects in the generated image, are they relevant with respect to the original image?}

The average rating for images from the experiment is $4.19$ out of a maximum rating of $5$, and it indicates that human users perceive the semantic fidelity of the generated images as a very high value, where semantic fidelity can be defined as the replication of semantics from transmitter to receiver. The experimentation is concerned with transmitting the images with zero semantic symbol errors in contrast to the traditional bit-error rate of the received image. In this experiment, we have measured the human perceived quality of the images transmitted with the developed semantic communication model. 

\section{Limitations of the Study}
The GAN that was  used in this setup is with pre-trained weights for the COCO dataset, and hence it
will fail if the semantic segmentation maps include edge distortions due to noise introduced in the physical channel. The GAN shall be trained with a higher number of object classes and with a much larger task-oriented dataset. Further, the used segmentation maps are the semantic feature extracted maps available in the coco data set. Future work will include a semantic extraction algorithm at the transmitter side. The experiment is only done for the AWGN channel; other channels, such as fading channels, may also be considered.  

\section{Conclusions}
In this work, we have introduced a semantic communication-based image transmission system for bandwidth-intensive mobile communication systems. Compared to the conventional image transmission systems, the designed semantic communications model could achieve a compression ratio of approximately $20$, which is a considerable improvement. The proposed semantic communication system transmits a semantic segmentation map extracted from the real image to the receiver over a noisy channel. We have filtered the salt and pepper noise at the receiver and input the resultant segmentation map as the condition to generate the desired image at the GAN.  

The study shows the impact of the physical channel noise on the semantic communication system and derives a threshold level of noise that the developed architecture can withhold. The developed architecture is resilient to a noise level of $2.1$ dB with bit error rates approximately equivalent to $0.4$ \%. We have derived an important finding from experiments conducted on the semantic communications model, where edge preservation is an essential function in SC. The channel coding in the physical channel will introduce edge distortions to the semantic segmentation maps, and it will cause the GAN to fail with the distorted semantic segmentation map. The purposed scheme outperforms JPEG compressed image transmission under low bit rate and low SNR scenarios. Although JPEG compression provided better quality for human perception in high SNR scenarios at the expense of a higher bit rate,  the intended application of this SC based communication systems is considered M2M communication. Hence,the proposed scheme outperforms traditional JPEG compressed transmission for all SNR values since it only consumes 5\% of the bandwidth JPEG used. The proposed semantic communication-based image transmission model opens up several research avenues in both practical applications, such as video transmission in AR/VR and video conferencing, and theoretical aspects. The proposed method enables the transmission of background information separately within a video frame. To achieve this, some modifications need to be made to the semantic extraction network by extracting the background and speakers separately. This could be viewed as a potential extension of the proposed architecture for video applications.

\section{Future Work}
Semantic communication is the next revolution in digital communication to cater to the ever-increasing need for bandwidth in terms of multimedia content. Although the basics are developed, their practical applications and real-world implementations are in their infancy. There are many research avenues to pursue in this direction, such as applying semantic communication theory to the transmission of video content in low bandwidth, unreliable mobile communication channels. The high data rate requirement, which needs huge bandwidths, can be effectively addressed via a suitably designed semantic communication link, thus paving the way for remote area access for many applications facilitating the digital revolution, which is currently missing in rural areas.  

Measuring the quality of the transmitted content in relation to human and machine perspectives has to be explored to develop new matrices to evaluate the quality of semantic communication. Traditional quality evaluation matrices, which were developed to bit-by-bit transmitted data are not compatible with measuring the semantic integrity of the transmitted data with semantic communication systems.
The effects of physical channel characteristics and different physical channels on semantic communication is also an area to be explored.  

\bibliographystyle{IEEEtran} 
\bibliography{Bib.bib}  

\begin{thebibliography}{10}
\providecommand{\url}[1]{#1}
\csname url@samestyle\endcsname
\providecommand{\newblock}{\relax}
\providecommand{\bibinfo}[2]{#2}
\providecommand{\BIBentrySTDinterwordspacing}{\spaceskip=0pt\relax}
\providecommand{\BIBentryALTinterwordstretchfactor}{4}
\providecommand{\BIBentryALTinterwordspacing}{\spaceskip=\fontdimen2\font plus
\BIBentryALTinterwordstretchfactor\fontdimen3\font minus
  \fontdimen4\font\relax}
\providecommand{\BIBforeignlanguage}[2]{{%
\expandafter\ifx\csname l@#1\endcsname\relax
\typeout{** WARNING: IEEEtran.bst: No hyphenation pattern has been}%
\typeout{** loaded for the language `#1'. Using the pattern for}%
\typeout{** the default language instead.}%
\else
\language=\csname l@#1\endcsname
\fi
#2}}
\providecommand{\BIBdecl}{\relax}
\BIBdecl

\bibitem{shannon_mathematical_1949}
C.~Shannon and W.~Weaver, \emph{The {Mathematical} {Theory} of
  {Communication}}.\hskip 1em plus 0.5em minus 0.4em\relax Urbana: University
  of Illinois Press, 1949.

\bibitem{goodfellow2020generative}
I.~Goodfellow, J.~Pouget-Abadie, M.~Mirza, B.~Xu, D.~Warde-Farley, S.~Ozair,
  A.~Courville, and Y.~Bengio, ``Generative adversarial networks,''
  \emph{Communications of the ACM}, vol.~63, no.~11, pp. 139--144, 2020.

\bibitem{Omijeh2016BPSK}
B.~O. Omijeh and T.~Oteheri, ``Binary phase shift keying digital modulation
  technique for noiseless and noisy transmission,'' \emph{Journal of Circuits
  and Systems}, vol.~5, p.~24, 2016.

\bibitem{BULL2021335}
\BIBentryALTinterwordspacing
D.~R. Bull and F.~Zhang, ``Chapter 10 - measuring and managing picture
  quality,'' in \emph{Intelligent Image and Video Compression (Second
  Edition)}, D.~R. Bull and F.~Zhang, Eds.\hskip 1em plus 0.5em minus
  0.4em\relax Oxford: Academic Press, 2021, pp. 335--384. [Online]. Available:
  \url{https://www.sciencedirect.com/science/article/pii/B9780128203538000190}
\BIBentrySTDinterwordspacing

\bibitem{6gWhitepaper2020}
\BIBentryALTinterwordspacing
S.~Ali, W.~Saad, N.~Rajatheva, K.~Chang, D.~Steinbach, B.~Sliwa, C.~Wietfeld,
  K.~Mei, H.~Shiri, H.-J. Zepernick, T.~M.~C. Chu, I.~Ahmad, J.~Huusko,
  J.~Suutala, S.~Bhadauria, V.~Bhatia, R.~Mitra, S.~Amuru, R.~Abbas, B.~Shao,
  M.~Capobianco, G.~Yu, M.~Claes, T.~Karvonen, M.~Chen, M.~Girnyk, and
  H.~Malik, ``{6G} white paper on machine learning in wireless communication
  networks,'' 2020. [Online]. Available: \url{https://arxiv.org/abs/2004.13875}
\BIBentrySTDinterwordspacing

\bibitem{semantic20216g}
\BIBentryALTinterwordspacing
E.~{Calvanese Strinati} and S.~Barbarossa, ``6g networks: Beyond shannon
  towards semantic and goal-oriented communications,'' \emph{Computer
  Networks}, vol. 190, p. 107930, 2021. [Online]. Available:
  \url{https://www.sciencedirect.com/science/article/pii/S1389128621000773}
\BIBentrySTDinterwordspacing

\bibitem{ZHANG202260}
\BIBentryALTinterwordspacing
P.~Zhang, W.~Xu, H.~Gao, K.~Niu, X.~Xu, X.~Qin, C.~Yuan, Z.~Qin, H.~Zhao,
  J.~Wei, and F.~Zhang, ``Toward wisdom-evolutionary and primitive-concise 6g:
  A new paradigm of semantic communication networks,'' \emph{Engineering},
  vol.~8, pp. 60--73, 2022. [Online]. Available:
  \url{https://www.sciencedirect.com/science/article/pii/S2095809921004513}
\BIBentrySTDinterwordspacing

\bibitem{wang2022transformer}
Y.~Wang, Z.~Gao, D.~Zheng, S.~Chen, D.~Gunduz, and H.~V. Poor,
  ``Transformer-empowered 6g intelligent networks: From massive mimo processing
  to semantic communication,'' \emph{IEEE Wireless Communications}, 2022.

\bibitem{EdgeSC2022}
P.~Dong, Q.~Wu, X.~Zhang, and G.~Ding, ``Edge semantic cognitive intelligence
  for 6g networks: Novel theoretical models, enabling framework, and typical
  applications,'' \emph{China Communications}, vol.~19, no.~8, pp. 1--14, 2022.

\bibitem{yang2022semantic}
W.~Yang, H.~Du, Z.~Q. Liew, W.~Y.~B. Lim, Z.~Xiong, D.~Niyato, X.~Chi, X.~S.
  Shen, and C.~Miao, ``Semantic communications for future internet:
  Fundamentals, applications, and challenges,'' \emph{IEEE Communications
  Surveys \& Tutorials}, 2022.

\bibitem{UBT2023}
S.~R. Pokhrel and J.~Choi, ``Understand-before-talk (ubt): A semantic
  communication approach to 6g networks,'' \emph{IEEE Transactions on Vehicular
  Technology}, vol.~72, no.~3, pp. 3544--3556, 2023.

\bibitem{sana2022learning}
M.~Sana and E.~C. Strinati, ``Learning semantics: An opportunity for effective
  6g communications,'' in \emph{2022 IEEE 19th Annual Consumer Communications
  \& Networking Conference (CCNC)}.\hskip 1em plus 0.5em minus 0.4em\relax
  IEEE, 2022, pp. 631--636.

\bibitem{deepJSCC2019}
E.~Bourtsoulatze, D.~B. Kurka, and D.~G{\"u}nd{\"u}z, ``Deep joint
  source-channel coding for wireless image transmission,'' \emph{IEEE
  Transactions on Cognitive Communications and Networking}, vol.~5, no.~3, pp.
  567--579, 2019.

\bibitem{Anelli}
V.~W. Anelli, Y.~Deldjoo, T.~Di~Noia, and D.~Malitesta, ``Deep learning-based
  adaptive image compression system for a real-world scenario,'' in \emph{2020
  IEEE Conference on Evolving and Adaptive Intelligent Systems (EAIS)}, 2020,
  pp. 1--8.

\bibitem{agustsson2019generative}
E.~Agustsson, M.~Tschannen, F.~Mentzer, R.~Timofte, and L.~V. Gool,
  ``Generative adversarial networks for extreme learned image compression,'' in
  \emph{Proceedings of the IEEE/CVF International Conference on Computer
  Vision}, 2019, pp. 221--231.

\bibitem{9065473Patel}
M.~I. Patel, S.~Suthar, and J.~Thakar, ``Survey on image compression using
  machine learning and deep learning,'' in \emph{2019 International Conference
  on Intelligent Computing and Control Systems (ICCS)}, 2019, pp. 1103--1105.

\bibitem{prakash2017semantic}
A.~Prakash, N.~Moran, S.~Garber, A.~DiLillo, and J.~Storer, ``Semantic
  perceptual image compression using deep convolution networks,'' in \emph{2017
  Data Compression Conference (DCC)}.\hskip 1em plus 0.5em minus 0.4em\relax
  IEEE, 2017, pp. 250--259.

\bibitem{9398576}
H.~Xie, Z.~Qin, G.~Y. Li, and B.-H. Juang, ``Deep learning enabled semantic
  communication systems,'' \emph{IEEE Transactions on Signal Processing},
  vol.~69, pp. 2663--2675, 2021.

\bibitem{ST}
Z.~Weng and Z.~Qin, ``Semantic communication systems for speech transmission,''
  \emph{IEEE Journal on Selected Areas in Communications}, vol.~39, no.~8, pp.
  2434--2444, 2021.

\bibitem{MLSC1}
H.~Zhang, S.~Shao, M.~Tao, X.~Bi, and K.~B. Letaief, ``Deep learning-enabled
  semantic communication systems with task-unaware transmitter and dynamic
  data,'' \emph{IEEE Journal on Selected Areas in Communications}, vol.~41,
  no.~1, pp. 170--185, 2022.

\bibitem{aldausari2022video}
N.~Aldausari, A.~Sowmya, N.~Marcus, and G.~Mohammadi, ``Video generative
  adversarial networks: a review,'' \emph{ACM Computing Surveys (CSUR)},
  vol.~55, no.~2, pp. 1--25, 2022.

\bibitem{pixTOpix2018}
P.~Isola, J.-Y. Zhu, T.~Zhou, and A.~A. Efros, ``Image-to-image translation
  with conditional adversarial networks,'' in \emph{2017 IEEE Conference on
  Computer Vision and Pattern Recognition (CVPR)}, 2017, pp. 5967--5976.

\bibitem{wang2018high}
T.-C. Wang, M.-Y. Liu, J.-Y. Zhu, A.~Tao, J.~Kautz, and B.~Catanzaro,
  ``High-resolution image synthesis and semantic manipulation with conditional
  gans,'' in \emph{Proceedings of the IEEE conference on computer vision and
  pattern recognition}, 2018, pp. 8798--8807.

\bibitem{deepImagecoding}
D.~Huang, X.~Tao, F.~Gao, and J.~Lu, ``Deep learning-based image semantic
  coding for semantic communications,'' in \emph{2021 IEEE Global
  Communications Conference (GLOBECOM)}, 2021, pp. 1--6.

\bibitem{sempyr}
A.~Shocher, Y.~Gandelsman, I.~Mosseri, M.~Yarom, M.~Irani, W.~T. Freeman, and
  T.~Dekel, ``Semantic pyramid for image generation,'' in \emph{2020 IEEE/CVF
  Conference on Computer Vision and Pattern Recognition (CVPR)}, 2020, pp.
  7455--7464.

\bibitem{coco}
T.-Y. Lin, M.~Maire, S.~Belongie, J.~Hays, P.~Perona, D.~Ramanan,
  P.~Doll{\'a}r, and C.~L. Zitnick, ``Microsoft coco: Common objects in
  context,'' in \emph{European conference on computer vision}.\hskip 1em plus
  0.5em minus 0.4em\relax Springer, 2014, pp. 740--755.

\bibitem{cocostuff}
H.~Caesar, J.~Uijlings, and V.~Ferrari, ``Coco-stuff: Thing and stuff classes
  in context,'' in \emph{2018 IEEE/CVF Conference on Computer Vision and
  Pattern Recognition}, 2018, pp. 1209--1218.

\bibitem{Semantic}
T.~Park, M.-Y. Liu, T.-C. Wang, and J.-Y. Zhu, ``Semantic image synthesis with
  spatially-adaptive normalization,'' in \emph{2019 IEEE/CVF Conference on
  Computer Vision and Pattern Recognition (CVPR)}, 2019, pp. 2332--2341.

\bibitem{CGAN}
\BIBentryALTinterwordspacing
M.~Mirza and S.~Osindero, ``Conditional generative adversarial nets,''
  \emph{CoRR}, vol. abs/1411.1784, 2014. [Online]. Available:
  \url{http://arxiv.org/abs/1411.1784}
\BIBentrySTDinterwordspacing

\bibitem{Arikan}
E.~Arikan, ``Channel polarization: A method for constructing capacity-achieving
  codes,'' in \emph{2008 IEEE International Symposium on Information Theory},
  2008, pp. 1173--1177.

\bibitem{Rezaei2022}
H.~Rezaei, V.~Ranasinghe, N.~Rajatheva, M.~Latva-aho, G.~Park, and O.-S. Park,
  ``Implementation of ultra-fast polar decoders,'' in \emph{2022 IEEE
  International Conference on Communications Workshops (ICC Workshops)}, 2022,
  pp. 235--241.

\bibitem{Rezaei20222}
H.~Rezaei, N.~Rajatheva, and M.~Latva-Aho, ``Low-latency multi-kernel polar
  decoders,'' \emph{IEEE Access}, vol.~10, pp. 119\,460--119\,474, 2022.

\bibitem{Rezaei20223}
\BIBentryALTinterwordspacing
H.~Rezaei, N.~Rajatheva, and M.~Latva-aho, ``A combinational multi-kernel
  decoder for polar codes,'' 2022. [Online]. Available:
  \url{https://arxiv.org/abs/2211.08778.}
\BIBentrySTDinterwordspacing

\bibitem{Combinational2023}
\BIBentryALTinterwordspacing
H.~Rezaei, N.~Rajatheva, and M.~Latva-Aho, ``High-throughput rate-flexible
  combinational decoders for multi-kernel polar codes,'' 2023. [Online].
  Available: \url{https://arxiv.org/abs/2301.10445}
\BIBentrySTDinterwordspacing

\bibitem{Ercan2017}
F.~Ercan, C.~Condo, and W.~J. Gross, ``{Reduced-Memory High-Throughput Fast-SSC
  Polar Code Decoder Architecture},'' \emph{IEEE Workshop on Signal Processing
  Systems (SiPS)}, pp. 1--6, 2017.

\bibitem{panchanathan1996jpeg}
S.~Panchanathan, N.~Gamaz, and A.~Jain, ``Jpeg based scalable image
  compression,'' \emph{Computer Communications}, vol.~19, no.~12, pp.
  1001--1013, 1996.

\bibitem{medianFilter}
\BIBentryALTinterwordspacing
MathWorks. medfilt2: 2-d median filtering. [Online]. Available:
  \url{https://uk.mathworks.com/help/images/ref/medfilt2.html}
\BIBentrySTDinterwordspacing

\bibitem{shi2021new}
G.~Shi, D.~Gao, X.~Song, J.~Chai, M.~Yang, X.~Xie, L.~Li, and X.~Li, ``A new
  communication paradigm: From bit accuracy to semantic fidelity,'' \emph{arXiv
  preprint arXiv:2101.12649}, 2021.

\end{thebibliography}


\begin{IEEEbiography}[{\includegraphics[width=1in,height=1.25in,clip,keepaspectratio]{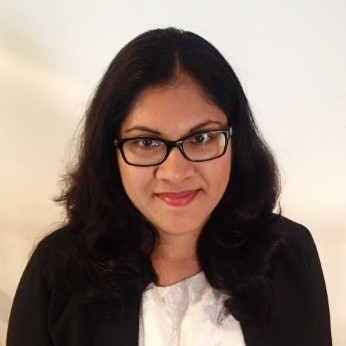}}]{Maheshi Lokumarambage} (Graduate Student Member, IEEE) received the B.Sc. (Hons.) degree specializing in computer engineering from the University of Peradeniya, Sri Lanka, in 2007, M.Sc. degree in computer science  from University of Moratuwa, Sri Lanka,in 2014 and MBA degree in Finance from University of Colombo, Sri Lanka, in 2012. She worked as an Engineer at Sri Lanka Telecom PLC, Sri Lanka, from 2007 to 2022. She is currently a doctoral researcher in the Dept. of Computer and Information Sciences, University of Strathclyde, UK. Her research interests include semantic communication, computer vision, and deep learning based resource optimization for video applications. 
\end{IEEEbiography}

\begin{IEEEbiography}[{\includegraphics[width=1in,height=1.25in,clip,keepaspectratio]{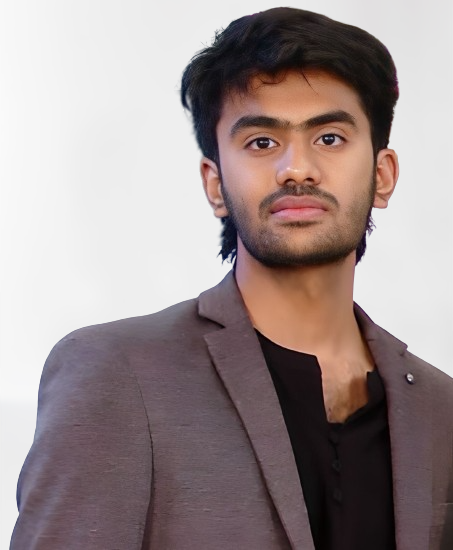}}]{Vishnu Sai Sankeerth Gowrisetty}
(Graduate Student Member, IEEE) received his Bachelor of Technology degree and Master of Technology degree in Electronics and Communications Engineering with a specialization in Signal Processing and Pattern Recognition from International Institute of Information Technology Bangalore, India in 2020. He worked on Localization of multi-labeled 3D Cryo Electron Tomograms using CNNs as part of research internship at Carnegie Mellon University, USA in 2019. He is currently a doctoral researcher in the Dept. of Computer and Information Sciences, University of Strathclyde, UK. His current research interests include Deep Learning based Video Codecs, Computer Vision and Automatic Speech Recognition.
\end{IEEEbiography}

\begin{IEEEbiography}[{\includegraphics[width=1in,height=1.25in,clip,keepaspectratio]{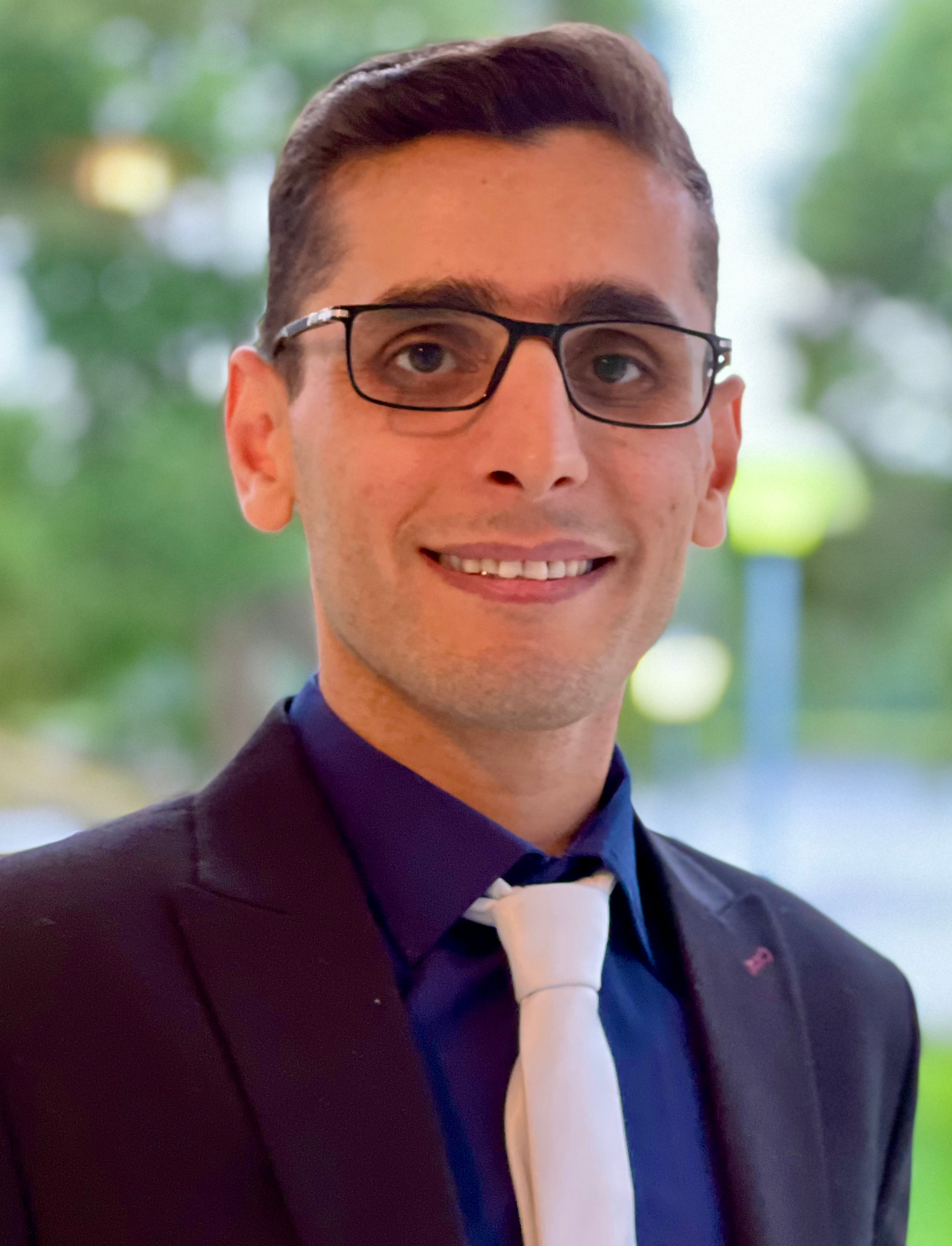}}]{Hossein Rezaei} (Graduate Student Member, IEEE) received his M.Sc degree in digital electronics from Iran University of Science and Technology, Tehran, Iran, in 2016. With over five years of experience as an FPGA/SoC designer in the industry, he has started his doctoral studies in Communications Engineering since 2020 at the University of Oulu, Oulu, Finland. He is also working as a senior SoC design engineer at Nokia, Oulu, Finland. His current research interests include design and implementation of error-correcting algorithms with a focus on polar codes, VLSI design for digital signal processing, semantic-based end-to-end transmission systems, and implementation of communication systems on embedded platform. 
\end{IEEEbiography}

\begin{IEEEbiography}[{\includegraphics[width=1in,height=1.25in,clip,keepaspectratio]{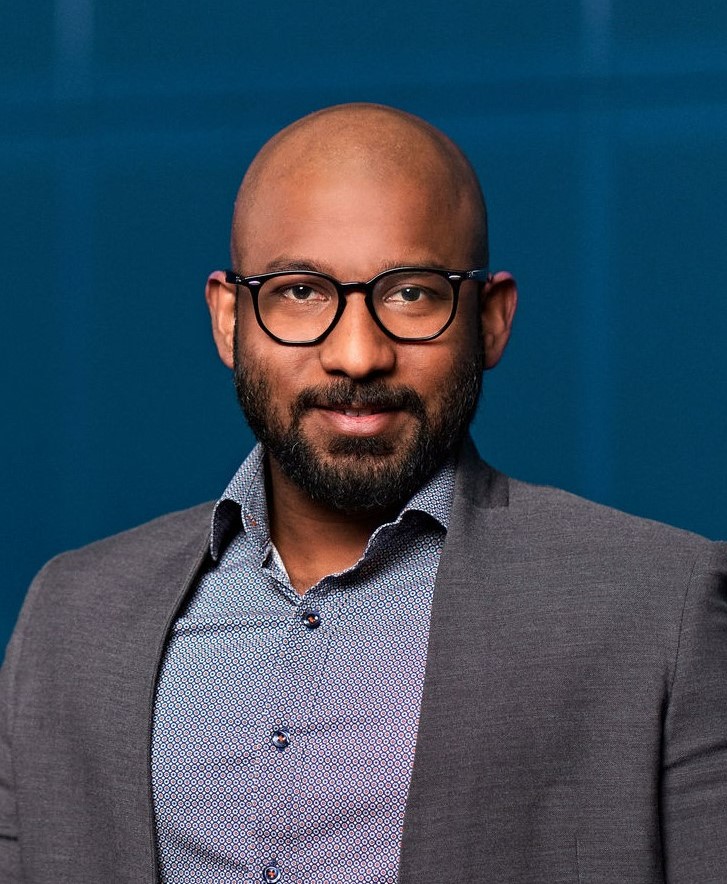}}]{Thushan Sivalingam} (Graduate Student Member, IEEE) received the B.Sc. (Hons.) degree specializing in Electrical and electronic engineering from the University of Peradeniya, Sri Lanka, in 2015, double degree M.Sc (tech.) in Wireless Communication from the University of Oulu, Finland, and the University of Peradeniya, Sri Lanka in 2019. He worked as a Planning Engineer-Access Network Planning at Dialog Axiata PLC, Sri Lanka, from 2015 to 2018. He is currently a doctoral researcher with the Centre for Wireless Communication, University of Oulu, Finland. His research interests include deep learning-based efficient and fast access methods for mMTC, grant-free NOMA, THz sensing, and semantic communications. 
\end{IEEEbiography}

\begin{IEEEbiography}[{\includegraphics[width=1in,height=1.25in,clip,keepaspectratio]{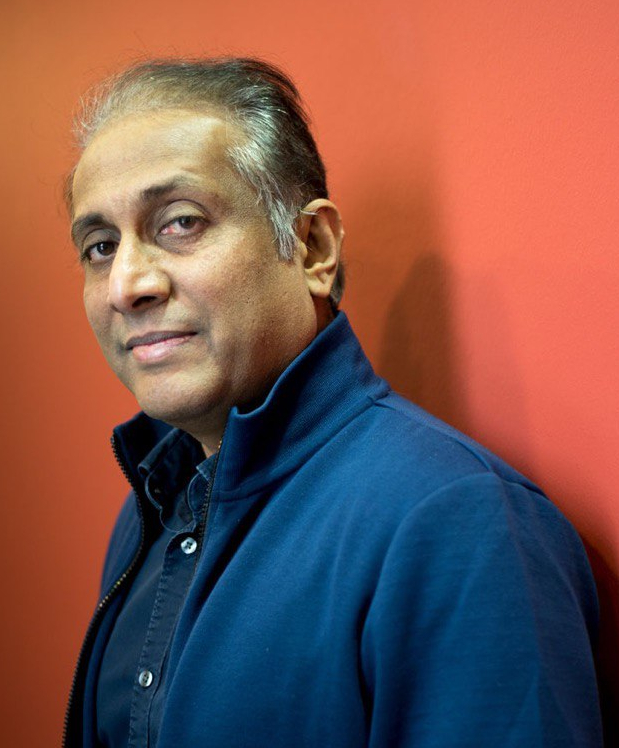}}]{Nandana Rajatheva}
(Senior Member, IEEE) received the B.Sc. (Hons.) degree in electronics and telecommunication engineering from the University of Moratuwa, Sri Lanka, in 1987, and the M.Sc. and Ph.D. degrees from the University of Manitoba, Winnipeg, MB, Canada, in 1991 and 1995, respectively. He is currently a Professor with the Centre for Wireless Communications, University of Oulu, Finland. During his graduate studies, he was a Canadian Commonwealth Scholar in Manitoba. From 1995 to 2010, he held a professor/associate professor positions with the University of Moratuwa and the Asian Institute of Technology, Thailand. He is currently leading the AI-driven Air Interface Design Task in Hexa-X EU Project. He has coauthored more than 200 referred articles published in journals and in conference proceedings. His research interests include physical layer in beyond 5G, machine learning for PHY and MAC, integrated sensing and communications as well as channel coding. 
\end{IEEEbiography}

\begin{IEEEbiography}[{\includegraphics[width=1in,height=1.25in,clip,keepaspectratio]{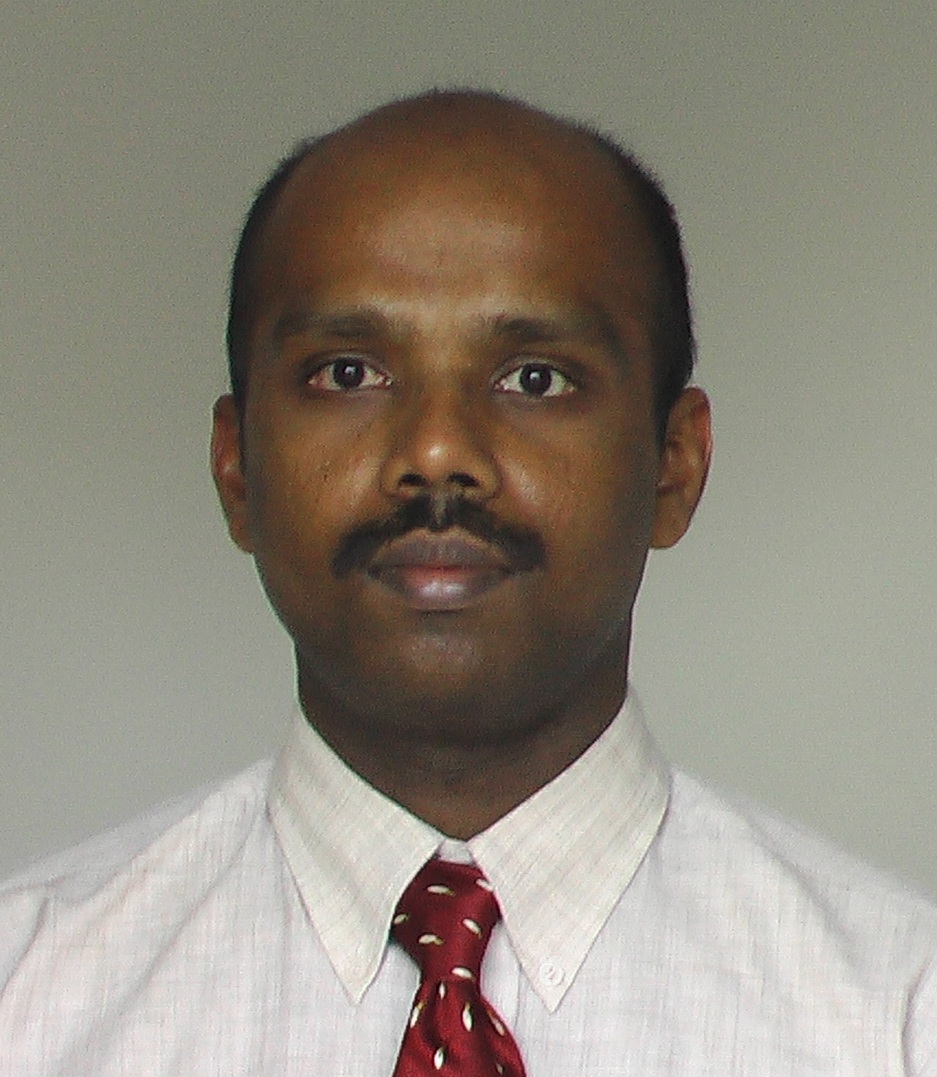}}]{Anil Fernando }
(Senior Member, IEEE) received the B.Sc. (Hons.) degree(First Class) in electronics and telecommunication engineering from the University of Moratuwa, Sri Lanka, in 1995, and the M.Sc. in Communications (Distinction) from the Asisan Institute of Technology, Bangkok, Thailand in 1997 and Ph.D. in Computer Science (Video Coding and Communications) from the University of Bristol, UK in 2001. He is a professor in Video Coding and Communications at the Department of Computer and Information Sciences, University of Strathclyde, UK. He leads the video coding and communication research team at Strathclyde. He has worked on major national and international multidisciplinary research projects and led most of them. He has published over 400 papers in international journals and conference proceedings and published a book on 3D video broadcasting. He has been working with all major EU broadcasters, BBC, and major European media companies/SMEs in the last decade in providing innovative media technologies for British and EU citizens. His main research interests are in Video coding and Communications, Machine Learning (ML) and Artificial Intelligence (AI), Semantic Communications, Signal Processing, Networking and Communications, Interactive Systems, Resource Optimizations in 6G, Distributed Technologies, Media Broadcasting, and Quality of Experience (QoE).
\end{IEEEbiography}

\end{document}